\documentclass[10pt,twocolumn]{article}

\newif\ifreview \reviewfalse

\usepackage[a4paper,top=2.2cm,bottom=2.4cm,left=1.9cm,right=1.9cm,columnsep=0.65cm,headsep=0.7cm,footskip=1cm]{geometry}
\usepackage[T1]{fontenc}
\IfFileExists{XCharter.sty}{\usepackage{XCharter}}{\usepackage{charter}}
\IfFileExists{sourcesanspro.sty}{\usepackage[scale=0.9]{sourcesanspro}}{\usepackage[scaled=0.88]{helvet}}
\usepackage{microtype}
\usepackage[table]{xcolor}
\usepackage{graphicx}
\usepackage{booktabs}
\usepackage{longtable}
\usepackage{pdflscape}
\usepackage{array}
\definecolor{rowshade}{gray}{0.93}
\newcolumntype{P}[1]{>{\raggedright\arraybackslash}p{#1}}
\usepackage{setspace}
\usepackage[super,sort&compress]{natbib}
\usepackage{titlesec}
\usepackage{needspace}
\usepackage{fancyhdr}
\usepackage{caption}
\usepackage{lineno}
\usepackage[hidelinks,pdftitle={The Value of System 1 Models in Medicine}]{hyperref}
\ifreview \linenumbers \else  \fi

\titleformat{\section}{\needspace{4\baselineskip}\sffamily\bfseries\small}{}{0pt}{\MakeUppercase}
\titlespacing*{\section}{0pt}{2.4ex plus .5ex minus .2ex}{0.9ex plus .1ex}
\titleformat{\subsection}[runin]{\normalfont\bfseries}{}{0pt}{}[.]
\titlespacing*{\subsection}{0pt}{1.4ex plus .3ex minus .2ex}{0.5em}

\DeclareCaptionLabelSeparator{bar}{\enspace|\enspace}
\fancypagestyle{first}{\fancyhf{}%
  \fancyfoot[R]{\sffamily\scriptsize\thepage}}

\newcommand{\Pval}{\textit{P}}
\newcommand{\COe}{CO\textsubscript{2}e}
\newcommand{\abshead}[1]{\textbf{#1}\enspace}

\begin{document}
\thispagestyle{first}

\twocolumn[{%
\begin{flushleft}
{\sffamily\scriptsize\MakeUppercase{Original research}\par}
\vspace{0.9em}
{\fontsize{24}{28}\selectfont The Value of System~1 Models in Medicine\par}
\vspace{1.1em}
{\sffamily\small\setstretch{1.25}
Xiufeng~Yang\textsuperscript{1,*},
Konstantinos~Vrettos\textsuperscript{2,*},
Zeineb~Ben~Chaaben\textsuperscript{1},
Fabian~Drexel\textsuperscript{1},
Jannik~L\"ubberstedt\textsuperscript{1},
Johannes~Moll\textsuperscript{1},
Anirudh~Narayanan\textsuperscript{1},
Frederike~Stumpf\textsuperscript{1},
Cosmin~I.~Bercea\textsuperscript{1},
Jiawen~Wang\textsuperscript{1},
Evangelia~E.~Vassalou\textsuperscript{3},
Krischan~Braitsch\textsuperscript{4},
Aymen~Meddeb\textsuperscript{5},
Sebastian~Ziegelmayer\textsuperscript{1},
Markus~Graf\textsuperscript{1},
Marcus~R.~Makowski\textsuperscript{1},
Lisa~Adams\textsuperscript{1},
Nassir~Navab\textsuperscript{6},
Michail~E.~Klontzas\textsuperscript{2,3,7,+},
Keno~Bressem\textsuperscript{1,8,9,10,+}\par}
\end{flushleft}
\vspace{0.6em}
\noindent\rule{\textwidth}{0.6pt}\par\vspace{0.9em}
\noindent\begin{minipage}[t]{0.25\textwidth}
{\sffamily\bfseries\small Abstract\par}
\end{minipage}%
\begin{minipage}[t]{0.75\textwidth}
\small\setstretch{1.05}\setlength{\parskip}{0.45em}\setlength{\parindent}{0pt}
\abshead{Background}Many clinical decisions are a choice among a few fixed options, taken quickly and in large numbers. Such decisions also need an uncertainty estimate to know when to defer to a clinician. Generative language models answer slowly in free text and state their confidence poorly. In contrast, System~1 models answer zero-shot with a probability for each answer. However, their performance across clinical tasks is unknown.

\abshead{Methods}Four System~1 models, two proprietary and two with open weights, were evaluated zero-shot on 86 public tasks from 14 clinical domains in 20 languages, spanning examination questions and medical imaging (radiography, CT and MRI). Tasks were assigned to six capability tiers before analysis. The primary outcome was the area under the receiver-operating-characteristic curve (AUROC). Secondary outcomes included calibration, language effects and comparisons with published results and generative models.

\abshead{Results}Across 179,111 cases, the mean AUROC of the best model per task was 0.94 for medical knowledge, 0.93 for text understanding, 0.83 for clinical judgement from text, 0.89 for image perception, 0.84 for 2D image findings and 0.69 for volumetric findings, with 0.54 for pulmonary embolism on CT. Imaging tasks scored 0.09 (95\% CI, 0.06 to 0.11) lower than text tasks (\Pval\,<\,0.001). The best model matched or exceeded the best published result on 13 of 26 text tasks and on 8 of 35 imaging tasks. Triage nurses outperformed the best model (accuracy, 0.85 vs.\ 0.56). One decision took 0.34 to 0.44 seconds and 24 to 81 joules on open-weight models run locally. Accuracy on identical questions fell by up to 39 percentage points from English to Yoruba. Generative models of the same family were 11 to 76 times slower, with 3 to 5 percentage points higher balanced accuracy.

\abshead{Conclusions}System~1 models decide quickly, cheaply and with usable confidence when the answer is stated in a text or recalled from medical knowledge, but not when it must be judged or read from an image.
\end{minipage}
\par\vspace{1.0em}
\noindent\rule{\textwidth}{0.3pt}\par\vspace{1.6em}
}]

{\let\thefootnote\relax
\footnotetext{\noindent\sffamily\scriptsize\setstretch{1.0}%
\textsuperscript{1}Department of Diagnostic and Interventional Radiology, School of Medicine and Health, Technical University of Munich, Munich, Germany.
\textsuperscript{2}Artificial Intelligence and Translational Imaging Lab, Department of Radiology, School of Medicine, University of Crete, Heraklion, Greece.
\textsuperscript{3}Department of Medical Imaging, University Hospital of Heraklion, Heraklion, Crete, Greece.
\textsuperscript{4}Department of Medicine III, School of Medicine and Health, Technical University of Munich, Munich, Germany.
\textsuperscript{5}Institute of Neuroradiology, Charit\'e -- Universit\"atsmedizin Berlin, Berlin, Germany.
\textsuperscript{6}Chair for Computer Aided Medical Procedures and Augmented Reality, School of Computation, Information and Technology, Technical University of Munich, Munich, Germany.
\textsuperscript{7}Division of Radiology, Department for Clinical Science, Intervention and Technology (CLINTEC), Karolinska Institutet, Stockholm, Sweden.
\textsuperscript{8}National Center for Tumor Diseases West, Essen, Germany.
\textsuperscript{9}Institute of Artificial Intelligence in Medicine, University Hospital Essen, Essen, Germany.
\textsuperscript{10}Department of Diagnostic and Interventional Radiology and Neuroradiology, University Hospital Essen, Essen, Germany.
\textsuperscript{*,+}These authors contributed equally.}}

\section*{Introduction}
Many clinical decisions represent choices among a few fixed options that must be made quickly and in bulk. A triage nurse assigns one of five acuity levels within minutes of arrival, a radiologist decides whether a CT pulmonary angiogram demonstrates embolism, and a coder selects one procedure code for each of thousands of cases. A model that makes such decisions must also state how certain it is, so that confident answers can be acted on and uncertain cases passed to a clinician.\citep{vancalster2019calibration,kompa2021second,dvijotham2023enhancing} Generative large language models perform well on a series of medical tasks,\citep{singhal2023large,nori2023capabilities} but they write each answer token by token, often following a reasoning trace, which takes several seconds per decision and uses far more energy than a task-specific model.\citep{luccioni2024power} The confidence they state in text is poorly calibrated and barely separates correct from incorrect answers.\citep{savage2025large,omar2025benchmarking} On the contrary, System~1 models receive a state, such as a report, an image or both, together with one or more typed questions, and return a probability for every answer option in a single forward pass without generating text. Their name refers to dual-process theory, which separates fast, intuitive System~1 judgement from slow, deliberate System~2 reasoning, the mode that generative models resemble when they reason in text.\citep{kahneman2011thinking,croskerry2009universal}

The open question is which clinical decisions such a model can be trusted with. The answer to a clinical case can be stated in a report, recalled from medical knowledge, judged from several findings, or seen in an image. Of 519 studies of large language models in health care, 44.5\% tested medical knowledge such as licensing-examination questions, and 5\% used real patient-care data.\citep{bedi2025testing} Where other tasks were tested, performance fell from more than 86\% on licensing-examination questions\citep{singhal2025toward} to lower accuracy on real clinical cases and conversations.\citep{hager2024evaluation,johri2025evaluation} On radiologic images, GPT-4V identified the imaging modality in all 515 images but detected abnormalities in 0\% to 90\%, depending on the finding,\citep{strotzer2024toward} and multimodal models scored lower on volumetric than on two-dimensional images.\citep{wu2026bilingual} If performance depends on where the answer lies, a single benchmark score misrepresents a model, a construct-validity problem known from general AI benchmarks.\citep{raji2021everything}

To evaluate the value of System~1 models in medicine, we tested four models on 86 tasks from 14 clinical domains, with the tasks ordered in six capability tiers from medical knowledge to volumetric findings. We measured which clinical decisions the models can take on, whether their probabilities are calibrated, how they perform outside English, and what their speed costs in time, money, energy and accuracy, against a generative model of the same family.

\section*{Methods}

\begin{figure*}[t]\centering
\includegraphics[width=\textwidth,height=0.65\textheight,keepaspectratio]{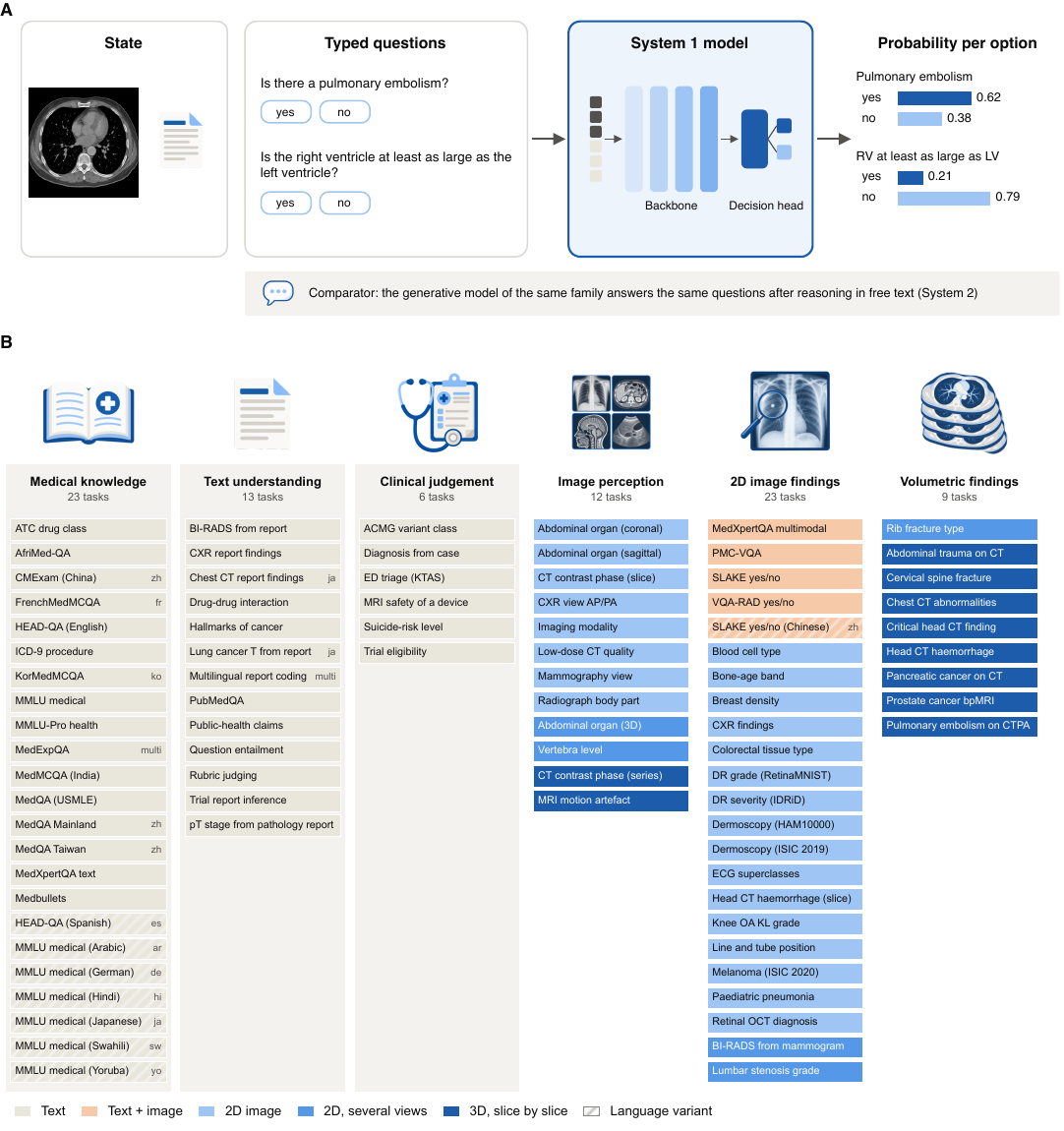}
\caption{\textbf{Study Design.} Panel A shows how a System~1 model answers. Text, one or more images, or both enters the model with typed questions, and the model returns a probability for each option in one forward pass. The generative model of the same family served as comparator on 32 tasks. Panel B shows the 86 tasks by capability tier and input type. Hatched tiles are language versions of another task and were analysed for language effects only.}
\label{fig:design}
\end{figure*}

\subsection*{Study Design and Tasks}
This cross-sectional evaluation used public, de-identified datasets under their respective licences. Candidate tasks came from a taxonomy of 87 clinical decision tasks in 18 domains and from further public datasets with a fixed answer format (Supplementary Appendix, Section~S1). A task was included when its data and reference labels were public and its licence allowed transfer to a model hosted on a commercial API. Eighty-six tasks from 14 clinical domains met these criteria. Table~S1 lists every task with the publication of its dataset. Eight of them are language versions of another task and were analysed only for language effects, which left 78 primary tasks (Figure~1). Each task used the official test split or a seeded random sample of up to 1,000 cases, in some tasks enriched for rare classes, and volumetric tasks used 81 to 436 studies. The tasks comprised 179,111 cases, with the reference standard released by the dataset authors.

Prior to result summarisation, each task was assigned to one of six capability tiers. Medical knowledge covered examination questions and terminology, text understanding covered information stated in a report, abstract or sentence, and clinical judgement from text covered decisions that integrate patient information, such as triage acuity or variant class. The three imaging tiers included image perception (acquisition, anatomy and image quality), two-dimensional (2D) image findings, and volumetric findings on CT or MRI.

\subsection*{System~1 Models}
Four System~1 models were evaluated through the OpenRouter decision interface, each in the latest version available in October 2026. Clef (Cloudflare, 27 billion parameters) and Clef-flash (Cloudflare, 9.5 billion parameters) have open weights, and GPT-6 Luna Decisions (OpenAI) and Jev~1.14 (TypeSafe AI) are proprietary. A request held a state of text and up to four images, and one or more typed questions with a yes-or-no answer, a choice of one of $n$ options, or an ordered score. The model returned a probability for every option. All questions of a case were sent in one request, zero-shot, with a one-sentence description of the input. Volumetric studies were read as axial slices at a fixed spacing, with one request per slice. The study-level probability was the maximum over slices for yes-or-no questions and the mean over slices for other questions. On four 2D imaging tasks, Jev~1.14 received higher-resolution versions of the same images since its interface accepts larger images. Every request was logged as sent.

\subsection*{Generative Comparison}
Clef and Clef-flash were compared with their generative backbones, Qwen3.8-27B\citep{qwen38} and Qwen3.5-9B,\citep{qwen3.5} and GPT-6 Luna Decisions with the generative GPT-6 Luna. The comparison covered 32 text and 2D imaging tasks. For each task, 100 randomly selected cases were sent to the generative model with the same state and questions as the System~1 model had received, including an instruction to choose one option per question, and an output schema restricted to the answer options. Default reasoning settings were kept. An answer that failed or could not be parsed counted as wrong.

\subsection*{Local Runs and Energy}
Clef and Clef-flash were additionally run locally from the released checkpoints, on a single NVIDIA RTX 5090 GPU. Local Clef's answers were compared with the hosted versions, which confirmed that the two agreed before the local runs were utilised. Energy was measured on a sub-sample of 42 tasks and it followed the AI Research Label of the CEDARS framework for reporting the environmental footprint of clinical AI.\citep{cedars} The measurements come from the GPU driver's cumulative energy counter through NVML (power usage effectiveness taken as 1.0 for a single workstation GPU). CodeCarbon (v3.3.1) ran alongside to convert energy to \COe{} at the local grid carbon intensity (Greece, 337\,g\,\COe{} per kWh). Energy is reported per decision and summarised as the median and IQR over tasks. Latency was measured per request. The providers of the hosted models (GPT-6 Luna Decisions, Jev~1.14) publish no per-request energy figures for them.

\subsection*{Reference Results}
For every task, published results on the same dataset were collected from the original dataset or challenge publication, from official leaderboards, and from a search of PubMed and the web for later work up to October 2026. Every value was checked against its source and classified as a task-specific model, a generative language model, a rule-based classifier, or clinicians and expert annotators. A published result was compared only when it used the same label set and metric on the matching questions. The highest such result per task is reported as the best published result.

\subsection*{Outcomes}
The primary outcome was the area under the receiver-operating-characteristic curve (AUROC), computed per question as a macro average of one-versus-rest curves for more than two options, and averaged over the questions of a task. Secondary outcomes were the balanced accuracy of the most probable option, expressed relative to chance as $(\mathrm{BA}-1/k)/(1-1/k)$ for $k$ classes so that 0 is chance and 1 is perfect, the top-label expected calibration error (ECE),\citep{guo2017calibration,vancalster2019calibration} selective accuracy after withholding the least confident answers,\citep{geifman2017selective} cost per 1,000 requests at provider prices, latency per request and energy per decision.

\subsection*{Statistical Analysis}
Any question that a model refused or failed was scored as wrong, with the label prevalence as its probability. Task-level 95\% confidence intervals (CIs) were estimated using 1,000 bootstrap resamples of cases. Task estimates were pooled within each tier and model using DerSimonian--Laird random-effects models. Two contrasts were specified in advance, imaging versus text tiers and a linear trend across the three imaging tiers, and both were estimated by random-effects meta-regression. AUROC was also summarised by question type. The accuracy of each language version on identical items was compared with English by paired bootstrap and the exact McNemar test. System~1 and generative models were compared by the mean task-level difference in balanced accuracy, with 95\% CIs from a task bootstrap. Secondary analyses were not adjusted for multiplicity. Analyses used Python 3.13, pandas 3.0, NumPy 2.5, SciPy 1.18 and scikit-learn 1.9. Code and all model outputs are available in the project repository at \url{https://github.com/twin-x-eu/system1-models-in-medicine}.

\section*{Results}

\subsection*{Tasks and Models}
The 86 tasks covered 20 languages and yielded 2.4~million model answers across 179,111 cases (Figure~1). GPT-6 Luna Decisions refused 4,008 case--question pairs (1.3\%), Clef and Clef-flash each failed on one pair, and Jev~1.14 answered every question. Refused and failed questions were scored as wrong. Jev~1.14 was the best model on 45 of 78 tasks. Results per task are given in Table~S2.

\begin{figure*}[t]\centering
\includegraphics[width=\textwidth,height=0.6\textheight,keepaspectratio]{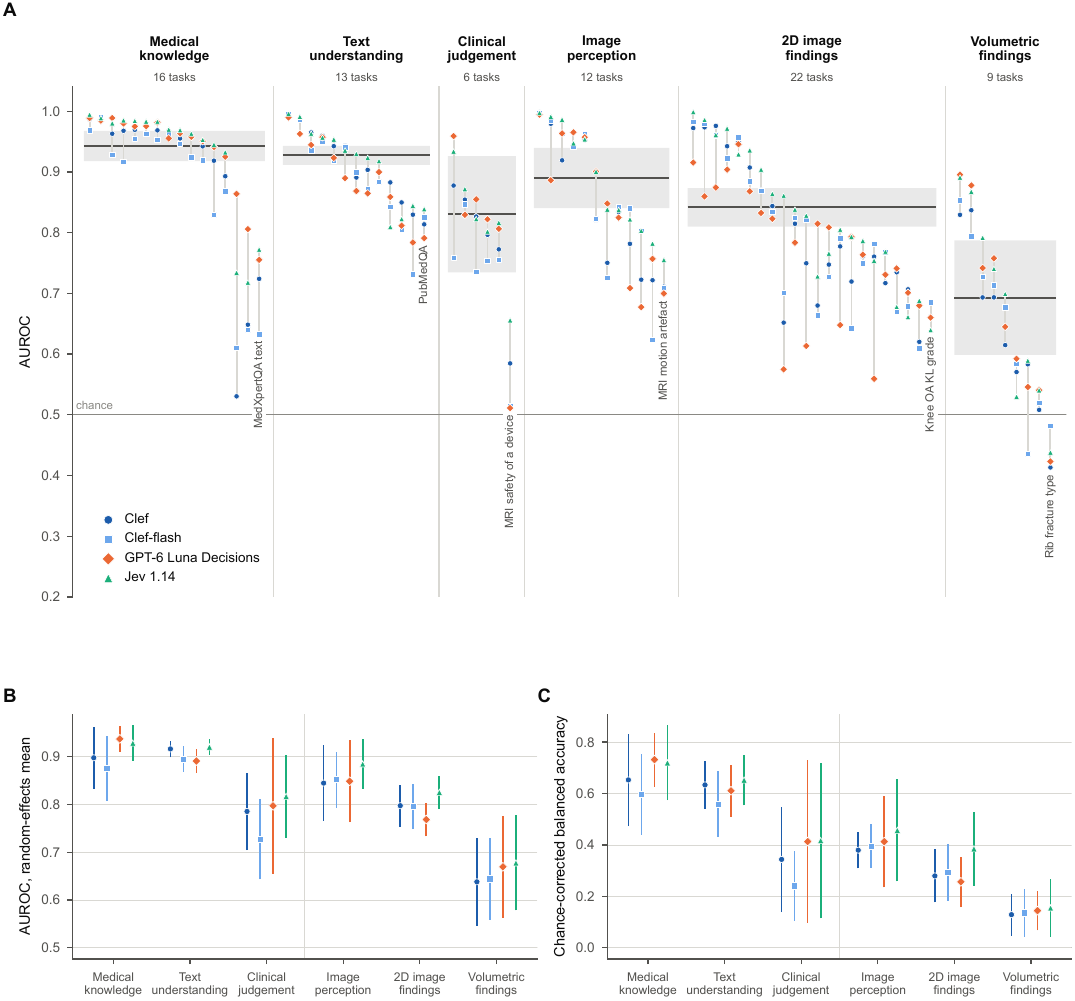}
\caption{\textbf{Capability Map.} Panel A shows the AUROC of each primary task (columns) for the four System~1 models, with tasks grouped by capability tier and sorted by the best model. Grey vertical lines span the four models. Horizontal bars and shaded bands show the random-effects mean AUROC of the best model per task with its 95\% CI, and the weakest task of each tier is labelled. Panels B and C show the random-effects mean per tier and model with 95\% CIs for AUROC and for chance-corrected balanced accuracy, where 0 is chance and 1 is perfect. Refused or failed questions were scored as wrong.}
\label{fig:capability}
\end{figure*}

\subsection*{Capability Gradient}
For the best model on each task, the random-effects mean AUROC was 0.94 (95\% CI, 0.92 to 0.97) for medical knowledge, 0.93 (95\% CI, 0.91 to 0.94) for text understanding, 0.89 (95\% CI, 0.84 to 0.94) for image perception, 0.84 (95\% CI, 0.81 to 0.87) for 2D image findings and 0.69 (95\% CI, 0.60 to 0.79) for volumetric findings (Figure~2). Imaging tasks scored 0.09 (95\% CI, 0.06 to 0.11) lower than text tasks (\Pval\,<\,0.001), and AUROC fell by 0.09 (95\% CI, 0.06 to 0.13) per tier from image perception to volumetric findings (\Pval\,<\,0.001). Both contrasts were significant for each of the four models (Figure~2B). Clinical judgement from text reached 0.83 (95\% CI, 0.73 to 0.93), lower than the other two text tiers. Chance-corrected balanced accuracy fell from 0.74 for medical knowledge to 0.20 for volumetric findings, and its trend across the imaging tiers was not significant (\Pval\,=\,0.10, Figure~2C).

The best model reached an AUROC of 0.99 on licensing-examination questions (MedQA)\citep{jin2021disease} and 1.00 on findings stated in chest radiograph reports. On volumetric imaging it reached 0.54 for pulmonary embolism,\citep{colak2021rsna} 0.59 for cervical spine fracture,\citep{lin2023rsna} 0.59 for clinically significant prostate cancer\citep{saha2024artificial} and 0.48 for rib fracture type, and four of nine volumetric tasks stayed below 0.60. Intracranial haemorrhage on head CT reached 0.90.\citep{chilamkurthy2018deep} On imaging tasks, questions that assigned the whole image to one of several options reached a median AUROC of 0.87, yes-or-no questions on the presence of a finding reached 0.74, and ordered grades 0.71. On text tasks, yes-or-no questions reached 0.99.

\begin{figure*}[t]\centering
\includegraphics[width=\textwidth,height=0.55\textheight,keepaspectratio]{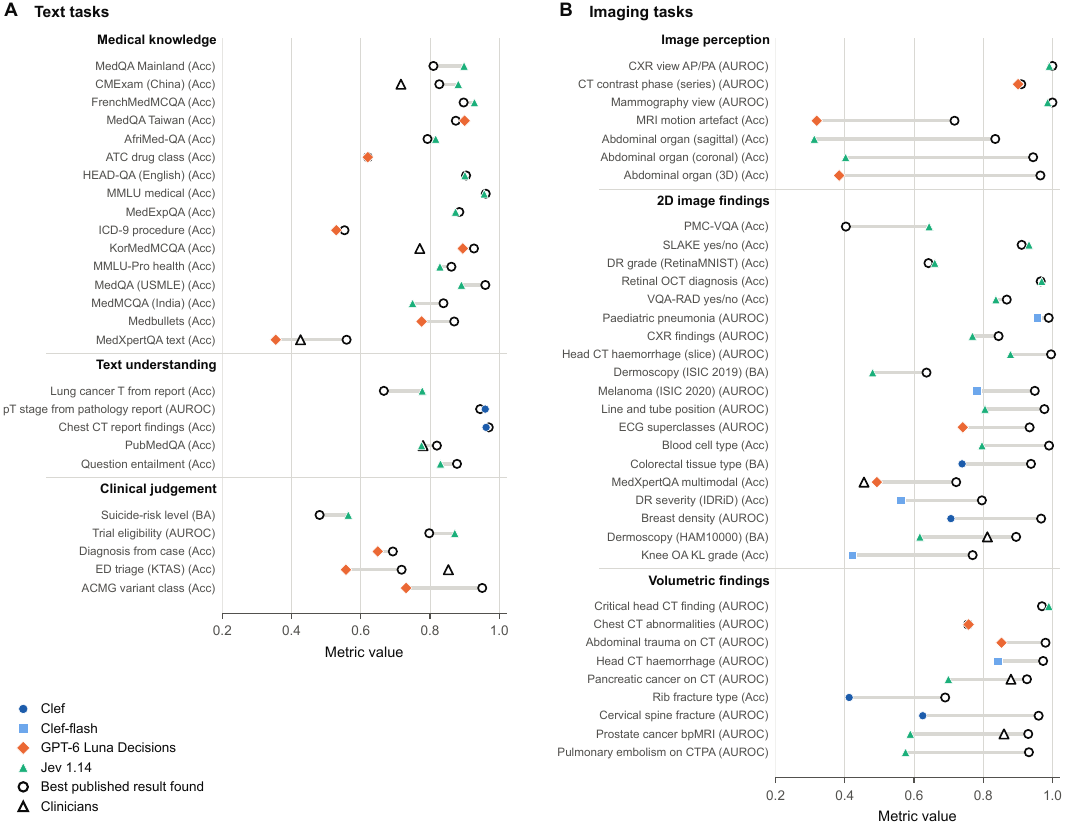}
\caption{\textbf{System~1 Models, Published Results and Clinicians.} Panel A shows text tasks and Panel B imaging tasks. For each task with a published result on the same dataset, label set and metric, the best of the four System~1 models (coloured marker) is shown with the best published result (open circle), which can be a task-specific model, a generative language model or a rule, and with clinicians or expert annotators (open triangle). Values of the System~1 models are computed on the questions that match the published result and can differ from the task-level AUROC. Acc denotes accuracy and BA balanced accuracy.}
\label{fig:published}
\end{figure*}

\subsection*{Comparison with Published Results and Clinicians}
A published result on the same dataset, label set and metric was available for 61 tasks (Figure~3). On text tasks, the best System~1 model matched (within 0.01) or exceeded the best published result in 13 of 26 tasks, with a median difference of $-$0.01. On imaging tasks it did so in 8 of 35 tasks, with a median difference of $-$0.17. Clinicians or expert annotators outperformed the best model in 6 of the 10 tasks for which such a comparison existed. Triage nurses assigned the emergency acuity level with an accuracy of 0.85, compared with 0.56 for the best model,\citep{moon2019triage} and a points-based rule classified genetic variants with an accuracy of 0.95, compared with 0.73.\citep{richards2015standards,tavtigian2020fitting} Zero-shot System~1 models are therefore competitive with purpose-built systems on text, but not on images or where a codified rule already exists.

\begin{figure*}[t]\centering
\includegraphics[width=\textwidth,height=0.42\textheight,keepaspectratio]{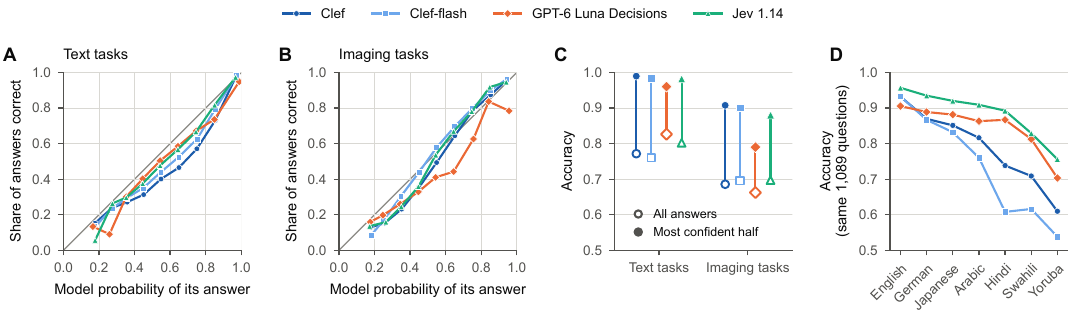}
\caption{\textbf{Calibration and Language.} Panels A and B show the share of correct answers against the probability the model gave its answer, pooled over text tasks in Panel A and over imaging tasks in Panel B. The diagonal marks perfect calibration. Panel C shows, for each model, the accuracy of all answers (open marker) and of its most confident half (filled marker) on text and imaging tasks. Panel D shows accuracy on the same 1,089 medical examination questions in seven languages.}
\label{fig:calibration}
\end{figure*}

\subsection*{Calibration and Selective Answering}
The median top-label ECE across models was 0.05 to 0.07 on text tasks and 0.11 to 0.17 on imaging tasks (Figure~4A and 4B). On imaging tasks, GPT-6 Luna Decisions was overconfident, and about 80\% of its most confident answers were correct. Withholding the less confident half of the answers raised accuracy on text tasks from between 0.76 and 0.83 to between 0.96 and 0.99, and on imaging tasks from between 0.66 and 0.70 to between 0.79 and 0.91 (Figure~4C).

\subsection*{Language}
On the same 1,089 medical examination questions, accuracy in every language version was lower than in English for every model (Figure~4D). The decline was smallest in German and largest in Yoruba, where it ranged from 20.1 (95\% CI, 17.6 to 22.5) percentage points for Jev~1.14 to 39.4 (95\% CI, 36.3 to 42.4) for Clef-flash, whose accuracy fell from 0.93 to 0.54. The performance measured in English therefore overstates what these models deliver in a low-resource language.

\begin{figure*}[t]\centering
\includegraphics[width=\textwidth,height=0.42\textheight,keepaspectratio]{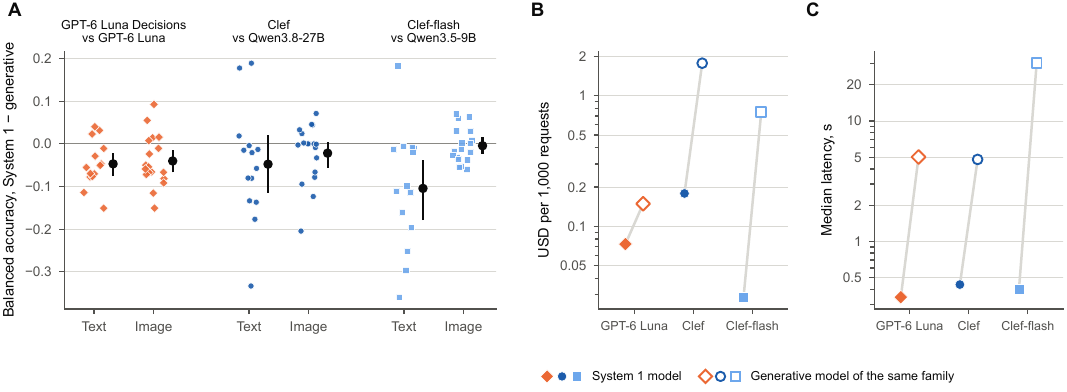}
\caption{\textbf{System~1 and Generative Models.} Panel A shows, for each task, the balanced accuracy of the System~1 model minus that of the generative model of the same family on identical cases, with the mean and 95\% CI (task bootstrap) for text and imaging tasks. The pairs are GPT-6 Luna Decisions and GPT-6 Luna, Clef and Qwen3.8-27B, and Clef-flash and Qwen3.5-9B. Panels B and C show the cost per 1,000 requests and the median latency per request on logarithmic scales, with filled markers for System~1 models and open markers for generative models.}
\label{fig:generative}
\end{figure*}

\subsection*{System~1 and Generative Models}
On identical cases, the balanced accuracy of each System~1 model was lower than that of the generative model of its family, by 4.3 (95\% CI, 2.4 to 6.3) percentage points for GPT-6 Luna Decisions, 3.3 (95\% CI, 0.0 to 6.5) for Clef and 4.8 (95\% CI, 1.0 to 8.5) for Clef-flash, over 32 tasks each (Figure~5A). The difference was smaller on imaging tasks, where Clef-flash and Qwen3.5-9B differed by 0.4 (95\% CI, $-$1.5 to 2.3) percentage points. Each System~1 model scored higher than its generative counterpart on 8 or 9 of the 32 tasks. Generative models required a median of 4.8 to 30.3 seconds per request, compared with 0.34 to 0.44 seconds for System~1 models, and incurred 2- to 22-fold higher costs (Figure~5B and 5C).

\begin{figure*}[t]\centering
\includegraphics[width=\textwidth,height=0.42\textheight,keepaspectratio]{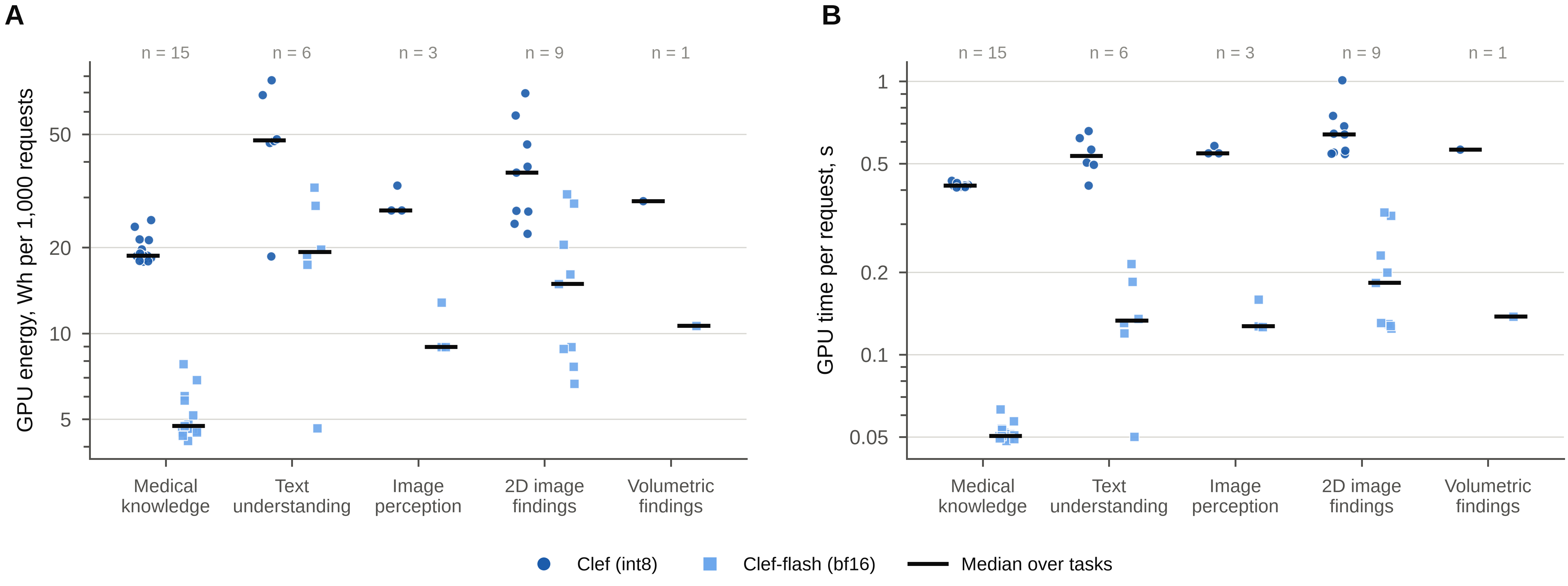}
\caption{\textbf{Energy of Locally Run System~1 Models.} Panel A shows the GPU energy per 1,000 requests and Panel B the GPU time per request of Clef (8-bit) and Clef-flash (bf16), run locally on one NVIDIA RTX 5090, for each primary task by capability tier. Energy is the whole-board GPU energy. Bars show the median over tasks, and n is the number of tasks per tier. No clinical-judgement task was run locally.}
\label{fig:energy}
\end{figure*}

\subsection*{Environmental Impact of System~1 Models}
Energy was measured directly on the two open-weight models, Clef and Clef-flash, both run locally on one RTX 5090 GPU over 42 of the tasks and 117,263 decisions each (Methods). The median energy of one decision was 23.5 joules for Clef-flash (IQR, 17.2 to 51.3) and 80.9 joules for Clef (IQR, 68.4 to 125.5). That is 6.5 to 22.5 watt-hours of GPU-board energy per 1,000 decisions and, including CPU and RAM and at the carbon intensity of the local grid, 2.8 to 12.1\,g\,\COe{} per 1,000 decisions (Figure~6). Across capability tiers, the median energy per 1,000 decisions was lowest for medical knowledge (4.8 watt-hours for Clef-flash and 19.2 for Clef) and highest for text understanding (19.3 and 47.6). One kilowatt-hour buys roughly 153,000 decisions from Clef-flash and 44,500 from Clef. Decisions that included an image cost 1.6 to 2.3 times as much energy as text-only decisions, and the length of the input mattered more than its type. On the three tasks with the longest documents, the energy of a decision rose to 2.1 to 5.0 times the median for that model. The two models together took 234,526 decisions for 4.3 kilowatt-hours of GPU-board energy. The energy of a single System~1 decision is therefore small enough that it is not the limiting factor in deploying one at scale.

\subsection*{Local versus API Inference}
Clef and Clef-flash have open weights and can run inside a hospital network, so patient data do not need to leave the institution. Run locally, they reproduced the hosted interface closely. On chest radiograph report findings, local Clef gave a mean AUROC of 0.99 equal to the version inferenced through the API, and agreed with it on 99.7\% of answers, while on visual question answering, the values were 0.947 against 0.942 and 97.2\% agreement.

\section*{Discussion}
System~1 models answered medical examination questions and identified findings described in radiology reports with high accuracy, with inference times below half a second and for a few joules per decision. On text-based tasks, predicted probabilities were well calibrated, closely reflecting observed accuracy. However, performance declined on tasks requiring the integration of multiple findings, such as assigning acuity levels in emergency triage. On imaging tasks, they accurately identified imaging modalities and anatomical regions but frequently failed to detect specific abnormalities. Performance on volumetric CT and MRI tasks was particularly limited, approaching chance levels in four of nine tasks.

For information extraction from clinical text, the case for zero-shot System~1 models is strong. The best System~1 model matched or exceeded the best published result on 13 of 26 text tasks without any training on the task, including results from models that had been fine-tuned for that task. For the T staging of lung cancer from CT reports, the best System~1 model reached an accuracy of 0.78, compared with 0.67 for fine-tuned Japanese BERT models.\citep{nakamura2023ntcir17} For the pathological T stage from pathology reports, it reached an AUROC of 0.96, compared with 0.94 for a fine-tuned Clinical-BigBird model.\citep{kefeli2024generalizable} A zero-shot System~1 model could therefore replace models that require task-specific fine-tuning. Each decision took less than half a second and 23.5 to 80.9 joules, with costs ranging from 0.02 to 0.16 USD per 1,000 requests. Withholding the 50\% of predictions with the lowest confidence increased accuracy on text-based tasks to 0.96 to 0.99. These findings suggest that System~1 models could support large-scale report labelling, information extraction, and literature screening, while referring uncertain cases to clinicians or slower, more computationally intensive models.\citep{geifman2017selective,dvijotham2023enhancing} Clef and Clef-flash have openly available weights and can be deployed within hospital networks.

Accuracy declined on tasks requiring clinical judgement, such as assigning triage acuity or classifying variants, even when all relevant information was provided in the input. System~1 models generate predictions in a single forward pass, relying on knowledge encoded in their weights. Unlike generative models that produce explicit reasoning traces,\citep{wei2023chainofthoughtpromptingelicitsreasoning} they lack intermediate reasoning steps through which multiple findings can be evaluated and integrated before reaching a decision. Across the five tasks requiring clinical judgement, the generative backbones of Clef and Clef-flash achieved accuracies 11 and 15 percentage points higher, respectively, compared with differences of 3 and 5 percentage points across all 32 evaluated tasks. Enabling System~1 models to generate reasoning traces before making predictions could help narrow this performance gap.

An additional limitation of most medical benchmarks is their predominantly English-language content. Similarly, most language models, including System~1 models, are trained primarily on English-language data.\citep{singh2025global,nimo2025afrimed} Reflecting this, the System~1 models were most accurate in English. On the same 1,089 examination questions, accuracy fell by 2 to 7 percentage points in German, a high-resource language, and by 20 to 39 percentage points in Yoruba, a low-resource language. Generative models lose accuracy in low-resource languages in a similar manner.\citep{nimo2025afrimed,alonso2024medexpqa}

Most System~1 models also accept images, but read medical images far less accurately than models trained for the task. The best System~1 model matched or exceeded the best published result in 8 of 35 imaging tasks, compared with 13 of 26 text tasks, with a median difference of $-$0.17 on imaging tasks. For clinically significant prostate cancer on MRI, it reached an AUROC of 0.59, compared with 0.86 for radiologists.\citep{saha2024artificial} Calibration was also worse on imaging tasks, and only about 80\% of the most confident answers of GPT-6 Luna Decisions were correct. Volumetric imaging was most challenging as the current interface of System~1 models accepts at most four images per request and no volumes. CT and MRI studies were therefore sent as single slices every 3 to 10\,mm, and a small finding can lie between two slices. The choice of pooling rule over slices did not change the results (Supplementary Appendix, Section~S6.3). Furthermore, radiographic images had to be downsampled, for example mammograms at 1,024 pixels on the long side, at which microcalcifications cover few pixels. Generative models also lose accuracy when volumes are given as image sequences,\citep{wu2026bilingual} which indicates a property of how vision-language models process images in general.

The open-weight decision models Clef and Clef-flash are based on the Qwen family of generative models, allowing a direct comparison of the two model types. The decision models and their backbones did not differ in accuracy, however the decision models were better calibrated, with a median expected calibration error of 0.12 for Clef-flash against 0.20 for its backbone, and 0.07 for Clef against 0.13. Against the generative backbone with reasoning, the System~1 models were 3 to 5 percentage points less accurate and 11 to 76 times faster.

This study has limitations. Models were used zero-shot with a one-sentence description of the input. Task-specific prompts and fine-tuning were not tested. A System~1 model trained on the records of 3.6 million patients from one health system, CS1-4B, was recently reported to outperform open decision models of up to 27 billion parameters on 13 of 14 tasks in held-out patients from that system.\citep{zhang2026generalpurposedecisionmodelclinical} Its weights have not yet been released, so it could not be included here. Public datasets can be part of the training data, which raises results on text tasks in particular. Larger frontier generative models were not tested and are likely more accurate on many tasks, but they are slower and more expensive than the generative models compared herein.

For predefined clinical decisions on text, in particular information extraction, System~1 models are ready for use, with fast, inexpensive and well-calibrated answers. They are as accurate as their backbones but better calibrated, and open weights let a hospital run them sustainably on its own data. However, they are currently limited in questions requiring clinical reasoning and in interpreting medical images, in particular CT and MRI volumes.

\section*{Acknowledgements}
\begingroup\footnotesize
\noindent\begin{minipage}[c]{1.6cm}
\includegraphics[width=\linewidth]{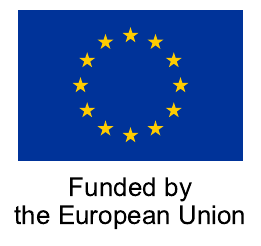}
\end{minipage}\hfill
\begin{minipage}[c]{\dimexpr\linewidth-2.0cm\relax}
Funded by the European Union under Horizon Europe grant agreement No~101287679 (TWIN-X). Views and opinions expressed are however those of the authors only and do not necessarily reflect those of the European Union or the European Health and Digital Executive Agency (HaDEA). Neither the European Union nor the granting authority can be held responsible for them.
\end{minipage}
\endgroup

{\footnotesize\setlength{\bibsep}{1pt plus 0.5pt}
\bibliographystyle{unsrtnat}
\bibliography{refs}}

@book{kahneman2011thinking,
  author = {Kahneman, Daniel},
  title = {{Thinking, Fast and Slow}},
  year = {2011},
  publisher = {Farrar, Straus and Giroux},
  address = {New York},
  isbn = {9780374275631}
}

@article{croskerry2009universal,
  author = {Croskerry, Pat},
  title = {A universal model of diagnostic reasoning},
  journal = {Acad Med},
  year = {2009},
  volume = {84},
  number = {8},
  pages = {1022--1028},
  doi = {10.1097/ACM.0b013e3181ace703},
  pmid = {19638766}
}

@article{singhal2023large,
  author = {Singhal, Karan and Azizi, Shekoofeh and Tu, Tao and Mahdavi, S Sara and Wei, Jason and Chung, Hyung Won and Scales, Nathan and Tanwani, Ajay and Cole-Lewis, Heather and Pfohl, Stephen and Payne, Perry and Seneviratne, Martin and Gamble, Paul and Kelly, Chris and Babiker, Abubakr and others},
  title = {Large language models encode clinical knowledge},
  journal = {Nature},
  year = {2023},
  volume = {620},
  number = {7972},
  pages = {172--180},
  doi = {10.1038/s41586-023-06291-2},
  pmid = {37438534}
}

@article{singhal2025toward,
  author = {Singhal, Karan and Tu, Tao and Gottweis, Juraj and Sayres, Rory and Wulczyn, Ellery and Amin, Mohamed and Hou, Le and Clark, Kevin and Pfohl, Stephen R and Cole-Lewis, Heather and Neal, Darlene and Rashid, Qazi Mamunur and Schaekermann, Mike and Wang, Amy and Dash, Dev and others},
  title = {Toward expert-level medical question answering with large language models},
  journal = {Nat Med},
  year = {2025},
  volume = {31},
  number = {3},
  pages = {943--950},
  doi = {10.1038/s41591-024-03423-7},
  pmid = {39779926}
}

@misc{nori2023capabilities,
  author = {Nori, Harsha and King, Nicholas and McKinney, Scott Mayer and Carignan, Dean and Horvitz, Eric},
  title = {Capabilities of {GPT-4} on medical challenge problems},
  year = {2023},
  doi = {10.48550/arXiv.2303.13375},
  eprint = {2303.13375},
  archiveprefix = {arXiv},
  note = {arXiv:2303.13375}
}

@article{jin2021disease,
  author = {Jin, Di and Pan, Eileen and Oufattole, Nassim and Weng, Wei-Hung and Fang, Hanyi and Szolovits, Peter},
  title = {What disease does this patient have? {A} large-scale open domain question answering dataset from medical exams},
  journal = {Appl Sci},
  year = {2021},
  volume = {11},
  number = {14},
  pages = {6421},
  doi = {10.3390/app11146421}
}

@inproceedings{zuo2025medxpertqa,
  author = {Zuo, Yuxin and Qu, Shang and Li, Yifei and Chen, Zhangren and Zhu, Xuekai and Hua, Ermo and Zhang, Kaiyan and Ding, Ning and Zhou, Bowen},
  title = {{MedXpertQA}: benchmarking expert-level medical reasoning and understanding},
  booktitle = {Proceedings of the 42nd International Conference on Machine Learning},
  series = {Proceedings of Machine Learning Research},
  year = {2025},
  volume = {267},
  pages = {80961--80990},
  doi = {10.48550/arXiv.2501.18362},
  url = {https://proceedings.mlr.press/v267/zuo25a.html},
  note = {arXiv:2501.18362}
}

@article{hager2024evaluation,
  author = {Hager, Paul and Jungmann, Friederike and Holland, Robbie and Bhagat, Kunal and Hubrecht, Inga and Knauer, Manuel and Vielhauer, Jakob and Makowski, Marcus and Braren, Rickmer and Kaissis, Georgios and Rueckert, Daniel},
  title = {Evaluation and mitigation of the limitations of large language models in clinical decision-making},
  journal = {Nat Med},
  year = {2024},
  volume = {30},
  number = {9},
  pages = {2613--2622},
  doi = {10.1038/s41591-024-03097-1},
  pmid = {38965432}
}

@article{johri2025evaluation,
  author = {Johri, Shreya and Jeong, Jaehwan and Tran, Benjamin A and Schlessinger, Daniel I and Wongvibulsin, Shannon and Barnes, Leandra A and Zhou, Hong-Yu and Cai, Zhuo Ran and Van Allen, Eliezer M and Kim, David and Daneshjou, Roxana and Rajpurkar, Pranav},
  title = {An evaluation framework for clinical use of large language models in patient interaction tasks},
  journal = {Nat Med},
  year = {2025},
  volume = {31},
  number = {1},
  pages = {77--86},
  doi = {10.1038/s41591-024-03328-5},
  pmid = {39747685}
}

@article{bedi2025testing,
  author = {Bedi, Suhana and Liu, Yutong and Orr-Ewing, Lucy and Dash, Dev and Koyejo, Sanmi and Callahan, Alison and Fries, Jason A and Wornow, Michael and Swaminathan, Akshay and Lehmann, Lisa Soleymani and Hong, Hyo Jung and Kashyap, Mehr and Chaurasia, Akash R and Shah, Nirav R and Singh, Karandeep and others},
  title = {Testing and evaluation of health care applications of large language models: a systematic review},
  journal = {JAMA},
  year = {2025},
  volume = {333},
  number = {4},
  pages = {319--328},
  doi = {10.1001/jama.2024.21700},
  pmid = {39405325}
}

@inproceedings{raji2021everything,
  author = {Raji, Inioluwa Deborah and Bender, Emily M. and Paullada, Amandalynne and Denton, Emily and Hanna, Alex},
  title = {{AI} and the everything in the whole wide world benchmark},
  booktitle = {Proceedings of the Neural Information Processing Systems Track on Datasets and Benchmarks},
  year = {2021},
  volume = {1},
  doi = {10.48550/arXiv.2111.15366},
  url = {https://datasets-benchmarks-proceedings.neurips.cc/paper/2021/hash/084b6fbb10729ed4da8c3d3f5a3ae7c9-Abstract-round2.html},
  note = {arXiv:2111.15366}
}

@article{strotzer2024toward,
  author = {Strotzer, Quirin D and Nieberle, Felix and Kupke, Laura S and Napodano, Gerardo and Muertz, Anna Katharina and Meiler, Stefanie and Einspieler, Ingo and Rennert, Janine and Strotzer, Michael and Wiesinger, Isabel and Wendl, Christina and Stroszczynski, Christian and Hamer, Okka W and Schicho, Andreas},
  title = {Toward foundation models in radiology? {Quantitative} assessment of {GPT-4V's} multimodal and multianatomic region capabilities},
  journal = {Radiology},
  year = {2024},
  volume = {313},
  number = {2},
  pages = {e240955},
  doi = {10.1148/radiol.240955},
  pmid = {39589253}
}

@article{wu2026bilingual,
  author = {Wu, Qingxia and Wu, Qingxia and Zhang, Peipei and Yi, Zhifeng and Shen, Yu and Bai, Yan and Tan, Hongna and Dong, Pei and Xue, Zhong and Roberts, Neil and Wang, Meiyun},
  title = {A bilingual benchmark for evaluating diagnostic performance of multimodal large language models in radiology ({RadM-Bench}): evaluation development and validation},
  journal = {J Med Internet Res},
  year = {2026},
  volume = {28},
  pages = {e92183},
  doi = {10.2196/92183},
  pmid = {42566748}
}

@article{hamamci2026generalist,
  author = {Hamamci, Ibrahim Ethem and Er, Sezgin and Wang, Chenyu and Almas, Furkan and Simsek, Ayse Gulnihan and Esirgun, Sevval Nil and Dogan, Irem and Durugol, Omer Faruk and Hou, Benjamin and Shit, Suprosanna and Dai, Weicheng and Xu, Murong and Reynaud, Hadrien and Dasdelen, Muhammed Furkan and Wittmann, Bastian and others},
  title = {Generalist foundation models from a multimodal dataset for {3D} computed tomography},
  journal = {Nat Biomed Eng},
  year = {2026},
  volume = {10},
  number = {8},
  pages = {1610--1628},
  doi = {10.1038/s41551-025-01599-y},
  pmid = {41680439}
}

@inproceedings{guo2017calibration,
  author = {Guo, Chuan and Pleiss, Geoff and Sun, Yu and Weinberger, Kilian Q.},
  title = {On calibration of modern neural networks},
  booktitle = {Proceedings of the 34th International Conference on Machine Learning},
  series = {Proceedings of Machine Learning Research},
  year = {2017},
  volume = {70},
  pages = {1321--1330},
  doi = {10.48550/arXiv.1706.04599},
  url = {https://proceedings.mlr.press/v70/guo17a.html},
  note = {arXiv:1706.04599}
}

@article{vancalster2019calibration,
  author = {Van Calster, Ben and McLernon, David J and van Smeden, Maarten and Wynants, Laure and Steyerberg, Ewout W and {Topic Group ‘Evaluating diagnostic tests and prediction models’ of the STRATOS initiative}},
  title = {Calibration: the {Achilles} heel of predictive analytics},
  journal = {BMC Med},
  year = {2019},
  volume = {17},
  number = {1},
  pages = {230},
  doi = {10.1186/s12916-019-1466-7},
  pmid = {31842878}
}

@article{kompa2021second,
  author = {Kompa, Benjamin and Snoek, Jasper and Beam, Andrew L},
  title = {Second opinion needed: communicating uncertainty in medical machine learning},
  journal = {NPJ Digit Med},
  year = {2021},
  volume = {4},
  number = {1},
  pages = {4},
  doi = {10.1038/s41746-020-00367-3},
  pmid = {33402680}
}

@inproceedings{geifman2017selective,
  author = {Geifman, Yonatan and El-Yaniv, Ran},
  title = {Selective classification for deep neural networks},
  booktitle = {Advances in Neural Information Processing Systems},
  year = {2017},
  volume = {30},
  doi = {10.48550/arXiv.1705.08500},
  url = {https://proceedings.neurips.cc/paper/2017/hash/4a8423d5e91fda00bb7e46540e2b0cf1-Abstract.html},
  note = {arXiv:1705.08500}
}

@article{dvijotham2023enhancing,
  author = {Dvijotham, Krishnamurthy Dj and Winkens, Jim and Barsbey, Melih and Ghaisas, Sumedh and Stanforth, Robert and Pawlowski, Nick and Strachan, Patricia and Ahmed, Zahra and Azizi, Shekoofeh and Bachrach, Yoram and Culp, Laura and Daswani, Mayank and Freyberg, Jan and Kelly, Christopher and Kiraly, Atilla and others},
  title = {Enhancing the reliability and accuracy of {AI-enabled} diagnosis via complementarity-driven deferral to clinicians},
  journal = {Nat Med},
  year = {2023},
  volume = {29},
  number = {7},
  pages = {1814--1820},
  doi = {10.1038/s41591-023-02437-x},
  pmid = {37460754}
}

@article{savage2025large,
  author = {Savage, Thomas and Wang, John and Gallo, Robert and Boukil, Abdessalem and Patel, Vishwesh and Safavi-Naini, Seyed Amir Ahmad and Soroush, Ali and Chen, Jonathan H},
  title = {Large language model uncertainty proxies: discrimination and calibration for medical diagnosis and treatment},
  journal = {J Am Med Inform Assoc},
  year = {2025},
  volume = {32},
  number = {1},
  pages = {139--149},
  doi = {10.1093/jamia/ocae254},
  pmid = {39396184}
}

@article{omar2025benchmarking,
  author = {Omar, Mahmud and Agbareia, Reem and Glicksberg, Benjamin S and Nadkarni, Girish N and Klang, Eyal},
  title = {Benchmarking the confidence of large language models in answering clinical questions: cross-sectional evaluation study},
  journal = {JMIR Med Inform},
  year = {2025},
  volume = {13},
  pages = {e66917},
  doi = {10.2196/66917},
  pmid = {40378406}
}

@inproceedings{singh2025global,
  author = {Singh, Shivalika and Romanou, Angelika and Fourrier, Clémentine and Adelani, David Ifeoluwa and Ngui, Jian Gang and Vila-Suero, Daniel and Limkonchotiwat, Peerat and Marchisio, Kelly and Leong, Wei Qi and Susanto, Yosephine and Ng, Raymond and Longpre, Shayne and Ruder, Sebastian and Ko, Wei-Yin and Bosselut, Antoine and others},
  title = {Global {MMLU}: understanding and addressing cultural and linguistic biases in multilingual evaluation},
  booktitle = {Proceedings of the 63rd Annual Meeting of the Association for Computational Linguistics (Volume 1: Long Papers)},
  year = {2025},
  pages = {18761--18799},
  doi = {10.18653/v1/2025.acl-long.919}
}

@inproceedings{nimo2025afrimed,
  author = {Nimo, Charles and Olatunji, Tobi and Owodunni, Abraham Toluwase and Abdullahi, Tassallah and Ayodele, Emmanuel and Sanni, Mardhiyah and Aka, Ezinwanne C. and Omofoye, Folafunmi and Yuehgoh, Foutse and Faniran, Timothy and Dossou, Bonaventure F. P. and Yekini, Moshood O. and Kemp, Jonas and Heller, Katherine A and Omeke, Jude Chidubem and others},
  title = {{AfriMed-QA}: a {Pan-African}, multi-specialty, medical question-answering benchmark dataset},
  booktitle = {Proceedings of the 63rd Annual Meeting of the Association for Computational Linguistics (Volume 1: Long Papers)},
  year = {2025},
  pages = {1948--1973},
  doi = {10.18653/v1/2025.acl-long.96}
}

@article{alonso2024medexpqa,
  author = {Alonso, Iñigo and Oronoz, Maite and Agerri, Rodrigo},
  title = {{MedExpQA}: multilingual benchmarking of large language models for medical question answering},
  journal = {Artif Intell Med},
  year = {2024},
  volume = {155},
  pages = {102938},
  doi = {10.1016/j.artmed.2024.102938},
  pmid = {39121544}
}

@inproceedings{luccioni2024power,
  author = {Luccioni, Sasha and Jernite, Yacine and Strubell, Emma},
  title = {Power hungry processing: watts driving the cost of {AI} deployment?},
  booktitle = {Proceedings of the 2024 ACM Conference on Fairness, Accountability, and Transparency},
  year = {2024},
  pages = {85--99},
  doi = {10.1145/3630106.3658542}
}

@article{moon2019triage,
  author = {Moon, Sun-Hee and Shim, Jae Lan and Park, Keun-Sook and Park, Chon-Suk},
  title = {Triage accuracy and causes of mistriage using the {Korean} {Triage} and {Acuity} {Scale}},
  journal = {PLoS One},
  year = {2019},
  volume = {14},
  number = {9},
  pages = {e0216972},
  doi = {10.1371/journal.pone.0216972},
  pmid = {31490937}
}

@article{richards2015standards,
  author = {Richards, Sue and Aziz, Nazneen and Bale, Sherri and Bick, David and Das, Soma and Gastier-Foster, Julie and Grody, Wayne W and Hegde, Madhuri and Lyon, Elaine and Spector, Elaine and Voelkerding, Karl and Rehm, Heidi L and {ACMG Laboratory Quality Assurance Committee}},
  title = {Standards and guidelines for the interpretation of sequence variants: a joint consensus recommendation of the {American} {College} of {Medical} {Genetics} and {Genomics} and the {Association} for {Molecular} {Pathology}},
  journal = {Genet Med},
  year = {2015},
  volume = {17},
  number = {5},
  pages = {405--424},
  doi = {10.1038/gim.2015.30},
  pmid = {25741868}
}

@article{tavtigian2020fitting,
  author = {Tavtigian, Sean V and Harrison, Steven M and Boucher, Kenneth M and Biesecker, Leslie G},
  title = {Fitting a naturally scaled point system to the {ACMG/AMP} variant classification guidelines},
  journal = {Hum Mutat},
  year = {2020},
  volume = {41},
  number = {10},
  pages = {1734--1737},
  doi = {10.1002/humu.24088},
  pmid = {32720330}
}

@article{rehm2015clingen,
  author = {Rehm, Heidi L and Berg, Jonathan S and Brooks, Lisa D and Bustamante, Carlos D and Evans, James P and Landrum, Melissa J and Ledbetter, David H and Maglott, Donna R and Martin, Christa Lese and Nussbaum, Robert L and Plon, Sharon E and Ramos, Erin M and Sherry, Stephen T and Watson, Michael S and {ClinGen}},
  title = {{ClinGen--the} {Clinical} {Genome} {Resource}},
  journal = {N Engl J Med},
  year = {2015},
  volume = {372},
  number = {23},
  pages = {2235--2242},
  doi = {10.1056/NEJMsr1406261},
  pmid = {26014595}
}

@inproceedings{hendrycks2021measuring,
  author = {Hendrycks, Dan and Burns, Collin and Basart, Steven and Zou, Andy and Mazeika, Mantas and Song, Dawn and Steinhardt, Jacob},
  title = {Measuring massive multitask language understanding},
  booktitle = {International Conference on Learning Representations (ICLR)},
  year = {2021},
  doi = {10.48550/arXiv.2009.03300},
  url = {https://iclr.cc/virtual/2021/poster/2962},
  note = {arXiv:2009.03300}
}

@inproceedings{jin2019pubmedqa,
  author = {Jin, Qiao and Dhingra, Bhuwan and Liu, Zhengping and Cohen, William and Lu, Xinghua},
  title = {{PubMedQA}: a dataset for biomedical research question answering},
  booktitle = {Proceedings of the 2019 Conference on Empirical Methods in Natural Language Processing and the 9th International Joint Conference on Natural Language Processing (EMNLP-IJCNLP)},
  year = {2019},
  pages = {2567--2577},
  doi = {10.18653/v1/D19-1259}
}

@article{chilamkurthy2018deep,
  author = {Chilamkurthy, Sasank and Ghosh, Rohit and Tanamala, Swetha and Biviji, Mustafa and Campeau, Norbert G and Venugopal, Vasantha Kumar and Mahajan, Vidur and Rao, Pooja and Warier, Prashant},
  title = {Deep learning algorithms for detection of critical findings in head {CT} scans: a retrospective study},
  journal = {Lancet},
  year = {2018},
  volume = {392},
  number = {10162},
  pages = {2388--2396},
  doi = {10.1016/S0140-6736(18)31645-3},
  pmid = {30318264}
}

@inproceedings{wang2017chestx,
  author = {Wang, Xiaosong and Peng, Yifan and Lu, Le and Lu, Zhiyong and Bagheri, Mohammadhadi and Summers, Ronald M.},
  title = {{ChestX-ray8}: hospital-scale chest {X-ray} database and benchmarks on weakly-supervised classification and localization of common thorax diseases},
  booktitle = {Proceedings of the IEEE Conference on Computer Vision and Pattern Recognition (CVPR)},
  year = {2017},
  pages = {3462--3471},
  doi = {10.1109/CVPR.2017.369}
}

@article{wagner2020ptb,
  author = {Wagner, Patrick and Strodthoff, Nils and Bousseljot, Ralf-Dieter and Kreiseler, Dieter and Lunze, Fatima I and Samek, Wojciech and Schaeffter, Tobias},
  title = {{PTB-XL}, a large publicly available electrocardiography dataset},
  journal = {Sci Data},
  year = {2020},
  volume = {7},
  number = {1},
  pages = {154},
  doi = {10.1038/s41597-020-0495-6},
  pmid = {32451379}
}

@article{tschandl2018ham10000,
  author = {Tschandl, Philipp and Rosendahl, Cliff and Kittler, Harald},
  title = {The {HAM10000} dataset, a large collection of multi-source dermatoscopic images of common pigmented skin lesions},
  journal = {Sci Data},
  year = {2018},
  volume = {5},
  pages = {180161},
  doi = {10.1038/sdata.2018.161},
  pmid = {30106392}
}

@article{yang2023medmnist,
  author = {Yang, Jiancheng and Shi, Rui and Wei, Donglai and Liu, Zequan and Zhao, Lin and Ke, Bilian and Pfister, Hanspeter and Ni, Bingbing},
  title = {{MedMNIST} v2 - {A} large-scale lightweight benchmark for {2D} and {3D} biomedical image classification},
  journal = {Sci Data},
  year = {2023},
  volume = {10},
  number = {1},
  pages = {41},
  doi = {10.1038/s41597-022-01721-8},
  pmid = {36658144}
}

@article{colak2021rsna,
  author = {Colak, Errol and Kitamura, Felipe C and Hobbs, Stephen B and Wu, Carol C and Lungren, Matthew P and Prevedello, Luciano M and Kalpathy-Cramer, Jayashree and Ball, Robyn L and Shih, George and Stein, Anouk and Halabi, Safwan S and Altinmakas, Emre and Law, Meng and Kumar, Parveen and Manzalawi, Karam A and others},
  title = {The {RSNA} {Pulmonary} {Embolism} {CT} dataset},
  journal = {Radiol Artif Intell},
  year = {2021},
  volume = {3},
  number = {2},
  pages = {e200254},
  doi = {10.1148/ryai.2021200254},
  pmid = {33937862}
}

@article{lin2023rsna,
  author = {Lin, Hui Ming and Colak, Errol and Richards, Tyler and Kitamura, Felipe C and Prevedello, Luciano M and Talbott, Jason and Ball, Robyn L and Gumeler, Ekim and Yeom, Kristen W and Hamghalam, Mohammad and Simpson, Amber L and Strika, Jasna and Bulja, Deniz and Angkurawaranon, Salita and Pérez-Lara, Almudena and others},
  title = {The {RSNA} {Cervical} {Spine} {Fracture} {CT} dataset},
  journal = {Radiol Artif Intell},
  year = {2023},
  volume = {5},
  number = {5},
  pages = {e230034},
  doi = {10.1148/ryai.230034},
  pmid = {37795143}
}

@article{rudie2024rsna,
  author = {Rudie, Jeffrey D and Lin, Hui-Ming and Ball, Robyn L and Jalal, Sabeena and Prevedello, Luciano M and Nicolaou, Savvas and Marinelli, Brett S and Flanders, Adam E and Magudia, Kirti and Shih, George and Davis, Melissa A and Mongan, John and Chang, Peter D and Berger, Ferco H and Hermans, Sebastiaan and others},
  title = {The {RSNA} {Abdominal} {Traumatic} {Injury} {CT} ({RATIC}) dataset},
  journal = {Radiol Artif Intell},
  year = {2024},
  volume = {6},
  number = {6},
  pages = {e240101},
  doi = {10.1148/ryai.240101},
  pmid = {39441109}
}

@article{saha2024artificial,
  author = {Saha, Anindo and Bosma, Joeran S and Twilt, Jasper J and van Ginneken, Bram and Bjartell, Anders and Padhani, Anwar R and Bonekamp, David and Villeirs, Geert and Salomon, Georg and Giannarini, Gianluca and Kalpathy-Cramer, Jayashree and Barentsz, Jelle and Maier-Hein, Klaus H and Rusu, Mirabela and Rouvière, Olivier and others},
  title = {Artificial intelligence and radiologists in prostate cancer detection on {MRI} ({PI-CAI}): an international, paired, non-inferiority, confirmatory study},
  journal = {Lancet Oncol},
  year = {2024},
  volume = {25},
  number = {7},
  pages = {879--887},
  doi = {10.1016/S1470-2045(24)00220-1},
  pmid = {38876123}
}

@article{alves2026artificial,
  author = {Alves, Natalia and Schuurmans, Megan and Rutkowski, Dawid and Saha, Anindo and Vendittelli, Pierpaolo and Obuchowski, Nancy and Liedenbaum, Marjolein H and Haldorsen, Ingfrid S and Molven, Anders and Yakar, Derya and Geerdink, Jeroen and van Koeverden, Sebastiaan and Riviere, Deniece M and Venderink, Wulphert and de Haas, Robbert and others},
  title = {Artificial intelligence and radiologists in pancreatic cancer detection using standard of care {CT} scans ({PANORAMA}): an international, paired, non-inferiority, confirmatory, observational study},
  journal = {Lancet Oncol},
  year = {2026},
  volume = {27},
  number = {1},
  pages = {116--124},
  doi = {10.1016/S1470-2045(25)00567-4},
  pmid = {41275871}
}

@article{acevedo2020dataset,
  author = {Acevedo, Andrea and Merino, Anna and Alférez, Santiago and Molina, Ángel and Boldú, Laura and Rodellar, José},
  title = {A dataset of microscopic peripheral blood cell images for development of automatic recognition systems},
  journal = {Data Brief},
  year = {2020},
  volume = {30},
  pages = {105474},
  doi = {10.1016/j.dib.2020.105474},
  pmid = {32346559}
}

@misc{arora2025healthbench,
  author = {Arora, Rahul K and Wei, Jason and Hicks, Rebecca Soskin and Bowman, Preston and Qui{\~n}onero-Candela, Joaquin and Tsimpourlas, Foivos and Sharman, Michael and Shah, Meghan and Vallone, Andrea and Beutel, Alex and Heidecke, Johannes and Singhal, Karan},
  title = {{HealthBench}: evaluating large language models towards improved human health},
  year = {2025},
  doi = {10.48550/arXiv.2505.08775},
  eprint = {2505.08775},
  archiveprefix = {arXiv},
  note = {arXiv:2505.08775}
}

@article{baker2016automatic,
  author = {Baker, Simon and Silins, Ilona and Guo, Yufan and Ali, Imran and H{\"o}gberg, Johan and Stenius, Ulla and Korhonen, Anna},
  title = {Automatic semantic classification of scientific literature according to the hallmarks of cancer},
  journal = {Bioinformatics},
  year = {2016},
  volume = {32},
  number = {3},
  pages = {432--440},
  doi = {10.1093/bioinformatics/btv585},
  pmid = {26454282}
}

@article{bartnik2024waw,
  author = {Bartnik, Krzysztof and Bartczak, Tomasz and Krzyzi{\'n}ski, Mateusz and Korzeniowski, Krzysztof and Lamparski, Krzysztof and W{\k{e}}grzyn, Piotr and Lam, Eric and Bartkowiak, Mateusz and Wr{\'o}blewski, Tadeusz and Mech, Katarzyna and Januszewicz, Magdalena and Biecek, Przemys{\l}aw},
  title = {{WAW-TACE}: a hepatocellular carcinoma multiphase {CT} dataset with segmentations, radiomics features, and clinical data},
  journal = {Radiol Artif Intell},
  year = {2024},
  volume = {6},
  number = {6},
  pages = {e240296},
  doi = {10.1148/ryai.240296},
  pmid = {39441110}
}

@inproceedings{benabacha2019overview,
  author = {Ben Abacha, Asma and Shivade, Chaitanya and Demner-Fushman, Dina},
  title = {Overview of the {MEDIQA} 2019 shared task on textual inference, question entailment and question answering},
  booktitle = {Proceedings of the 18th BioNLP Workshop and Shared Task},
  year = {2019},
  pages = {370--379},
  doi = {10.18653/v1/W19-5039},
  url = {https://aclanthology.org/W19-5039/}
}

@article{bilic2023liver,
  author = {Bilic, Patrick and Christ, Patrick and Li, Hongwei Bran and Vorontsov, Eugene and Ben-Cohen, Avi and Kaissis, Georgios and Szeskin, Adi and Jacobs, Colin and Humpire Mamani, Gabriel Efrain and Chartrand, Gabriel and Loh{\"o}fer, Fabian and Holch, Julian Walter and Sommer, Wieland and Hofmann, Felix and Hostettler, Alexandre and others},
  title = {The {Liver Tumor Segmentation Benchmark} ({LiTS})},
  journal = {Med Image Anal},
  year = {2023},
  volume = {84},
  pages = {102680},
  doi = {10.1016/j.media.2022.102680},
  pmid = {36481607}
}

@misc{chen2018knee,
  author = {Chen, Pingjun},
  title = {Knee osteoarthritis severity grading dataset},
  year = {2018},
  publisher = {Mendeley Data},
  version = {1},
  doi = {10.17632/56rmx5bjcr.1},
  note = {Data set}
}

@article{chen2019fully,
  author = {Chen, Pingjun and Gao, Linlin and Shi, Xiaoshuang and Allen, Kyle and Yang, Lin},
  title = {Fully automatic knee osteoarthritis severity grading using deep neural networks with a novel ordinal loss},
  journal = {Comput Med Imaging Graph},
  year = {2019},
  volume = {75},
  pages = {84--92},
  doi = {10.1016/j.compmedimag.2019.06.002},
  pmid = {31238184}
}

@inproceedings{chen2025benchmarking,
  author = {Chen, Hanjie and Fang, Zhouxiang and Singla, Yash and Dredze, Mark},
  title = {Benchmarking large language models on answering and explaining challenging medical questions},
  booktitle = {Proceedings of the 2025 Conference of the Nations of the Americas Chapter of the Association for Computational Linguistics: Human Language Technologies (Volume 1: Long Papers)},
  year = {2025},
  pages = {3563--3599},
  doi = {10.18653/v1/2025.naacl-long.182}
}

@inproceedings{codella2018skin,
  author = {Codella, Noel C F and Gutman, David and Celebi, M Emre and Helba, Brian and Marchetti, Michael A and Dusza, Stephen W and Kalloo, Aadi and Liopyris, Konstantinos and Mishra, Nabin and Kittler, Harald and Halpern, Allan},
  title = {Skin lesion analysis toward melanoma detection: a challenge at the 2017 {International} {Symposium} on {Biomedical} {Imaging} ({ISBI}), hosted by the {International} {Skin} {Imaging} {Collaboration} ({ISIC})},
  booktitle = {2018 IEEE 15th International Symposium on Biomedical Imaging (ISBI 2018)},
  year = {2018},
  pages = {168--172},
  doi = {10.1109/ISBI.2018.8363547},
  note = {arXiv:1710.05006}
}

@misc{codella2019skin,
  author = {Codella, Noel and Rotemberg, Veronica and Tschandl, Philipp and Celebi, M Emre and Dusza, Stephen and Gutman, David and Helba, Brian and Kalloo, Aadi and Liopyris, Konstantinos and Marchetti, Michael and Kittler, Harald and Halpern, Allan},
  title = {Skin lesion analysis toward melanoma detection 2018: a challenge hosted by the {International} {Skin} {Imaging} {Collaboration} ({ISIC})},
  year = {2019},
  doi = {10.48550/arXiv.1902.03368},
  eprint = {1902.03368},
  archiveprefix = {arXiv},
  note = {arXiv:1902.03368}
}

@article{demnerfushman2016preparing,
  author = {Demner-Fushman, Dina and Kohli, Marc D and Rosenman, Marc B and Shooshan, Sonya E and Rodriguez, Laritza and Antani, Sameer and Thoma, George R and McDonald, Clement J},
  title = {Preparing a collection of radiology examinations for distribution and retrieval},
  journal = {J Am Med Inform Assoc},
  year = {2016},
  volume = {23},
  number = {2},
  pages = {304--310},
  doi = {10.1093/jamia/ocv080},
  pmid = {26133894}
}

@misc{fda2026accessgudid,
  author = {{U.S. Food and Drug Administration}},
  title = {{AccessGUDID}: {Global Unique Device Identification Database (GUDID)}},
  year = {2026},
  url = {https://accessgudid.nlm.nih.gov/},
  note = {Accessed October 2026}
}

@article{flanders2020construction,
  author = {Flanders, Adam E and Prevedello, Luciano M and Shih, George and Halabi, Safwan S and Kalpathy-Cramer, Jayashree and Ball, Robyn and Mongan, John T and Stein, Anouk and Kitamura, Felipe C and Lungren, Matthew P and Choudhary, Gagandeep and Cala, Lesley and Coelho, Luiz and Mogensen, Monique and Morón, Fanny and others},
  title = {Construction of a machine learning dataset through collaboration: the {RSNA} 2019 {Brain} {CT} {Hemorrhage} {Challenge}},
  journal = {Radiol Artif Intell},
  year = {2020},
  volume = {2},
  number = {3},
  pages = {e190211},
  doi = {10.1148/ryai.2020190211},
  pmid = {33937827}
}

@inproceedings{gaur2019knowledge,
  author = {Gaur, Manas and Alambo, Amanuel and Sain, Joy Prakash and Kursuncu, Ugur and Thirunarayan, Krishnaprasad and Kavuluru, Ramakanth and Sheth, Amit and Welton, Randy and Pathak, Jyotishman},
  title = {Knowledge-aware assessment of severity of suicide risk for early intervention},
  booktitle = {The World Wide Web Conference (WWW '19)},
  year = {2019},
  pages = {514--525},
  doi = {10.1145/3308558.3313698}
}

@article{halabi2019rsna,
  author = {Halabi, Safwan S and Prevedello, Luciano M and Kalpathy-Cramer, Jayashree and Mamonov, Artem B and Bilbily, Alexander and Cicero, Mark and Pan, Ian and Pereira, Lucas Ara{\'u}jo and Sousa, Rafael Teixeira and Abdala, Nitamar and Kitamura, Felipe Campos and Thodberg, Hans H and Chen, Leon and Shih, George and Andriole, Katherine and others},
  title = {The {RSNA} pediatric bone age machine learning challenge},
  journal = {Radiology},
  year = {2019},
  volume = {290},
  number = {2},
  pages = {498--503},
  doi = {10.1148/radiol.2018180736},
  pmid = {30480490}
}

@article{hernandezperez2024bcn20000,
  author = {Hernández-Pérez, Carlos and Combalia, Marc and Podlipnik, Sebastian and Codella, Noel C F and Rotemberg, Veronica and Halpern, Allan C and Reiter, Ofer and Carrera, Cristina and Barreiro, Alicia and Helba, Brian and Puig, Susana and Vilaplana, Veronica and Malvehy, Josep},
  title = {{BCN20000}: dermoscopic lesions in the wild},
  journal = {Sci Data},
  year = {2024},
  volume = {11},
  number = {1},
  pages = {641},
  doi = {10.1038/s41597-024-03387-w},
  pmid = {38886204}
}

@article{herrerozazo2013ddi,
  author = {Herrero-Zazo, Mar{\'i}a and Segura-Bedmar, Isabel and Mart{\'i}nez, Paloma and Declerck, Thierry},
  title = {The {DDI} corpus: an annotated corpus with pharmacological substances and drug-drug interactions},
  journal = {J Biomed Inform},
  year = {2013},
  volume = {46},
  number = {5},
  pages = {914--920},
  doi = {10.1016/j.jbi.2013.07.011},
  pmid = {23906817}
}

@article{jin2020deep,
  author = {Jin, Liang and Yang, Jiancheng and Kuang, Kaiming and Ni, Bingbing and Gao, Yiyi and Sun, Yingli and Gao, Pan and Ma, Weiling and Tan, Mingyu and Kang, Hui and Chen, Jiajun and Li, Ming},
  title = {Deep-learning-assisted detection and segmentation of rib fractures from {CT} scans: development and validation of {FracNet}},
  journal = {EBioMedicine},
  year = {2020},
  volume = {62},
  pages = {103106},
  doi = {10.1016/j.ebiom.2020.103106},
  pmid = {33186809}
}

@inproceedings{jullien2024semeval,
  author = {Jullien, Mael and Valentino, Marco and Freitas, Andr{\'e}},
  title = {{SemEval-2024} Task 2: safe biomedical natural language inference for clinical trials},
  booktitle = {Proceedings of the 18th International Workshop on Semantic Evaluation (SemEval-2024)},
  year = {2024},
  pages = {1947--1962},
  doi = {10.18653/v1/2024.semeval-1.271},
  url = {https://aclanthology.org/2024.semeval-1.271/}
}

@article{kather2019predicting,
  author = {Kather, Jakob Nikolas and Krisam, Johannes and Charoentong, Pornpimol and Luedde, Tom and Herpel, Esther and Weis, Cleo-Aron and Gaiser, Timo and Marx, Alexander and Valous, Nektarios A and Ferber, Dyke and Jansen, Lina and Reyes-Aldasoro, Constantino Carlos and Zörnig, Inka and Jäger, Dirk and Brenner, Hermann and others},
  title = {Predicting survival from colorectal cancer histology slides using deep learning: a retrospective multicenter study},
  journal = {PLoS Med},
  year = {2019},
  volume = {16},
  number = {1},
  pages = {e1002730},
  doi = {10.1371/journal.pmed.1002730},
  pmid = {30677016}
}

@article{kefeli2024generalizable,
  author = {Kefeli, Jenna and Berkowitz, Jacob and Acitores Cortina, Jose M and Tsang, Kevin K and Tatonetti, Nicholas P},
  title = {Generalizable and automated classification of {TNM} stage from pathology reports with external validation},
  journal = {Nat Commun},
  year = {2024},
  volume = {15},
  number = {1},
  pages = {8916},
  doi = {10.1038/s41467-024-53190-9},
  pmid = {39414770}
}

@article{kefeli2024tcga,
  author = {Kefeli, Jenna and Tatonetti, Nicholas},
  title = {{TCGA-Reports}: a machine-readable pathology report resource for benchmarking text-based {AI} models},
  journal = {Patterns (N Y)},
  year = {2024},
  volume = {5},
  number = {3},
  pages = {100933},
  doi = {10.1016/j.patter.2024.100933},
  pmid = {38487800}
}

@article{kermany2018identifying,
  author = {Kermany, Daniel S and Goldbaum, Michael and Cai, Wenjia and Valentim, Carolina C S and Liang, Huiying and Baxter, Sally L and McKeown, Alex and Yang, Ge and Wu, Xiaokang and Yan, Fangbing and Dong, Justin and Prasadha, Made K and Pei, Jacqueline and Ting, Magdalene Y L and Zhu, Jie and others},
  title = {Identifying medical diagnoses and treatable diseases by image-based deep learning},
  journal = {Cell},
  year = {2018},
  volume = {172},
  number = {5},
  pages = {1122--1131.e9},
  doi = {10.1016/j.cell.2018.02.010},
  pmid = {29474911}
}

@inproceedings{kotonya2020explainable,
  author = {Kotonya, Neema and Toni, Francesca},
  title = {Explainable automated fact-checking for public health claims},
  booktitle = {Proceedings of the 2020 Conference on Empirical Methods in Natural Language Processing (EMNLP)},
  year = {2020},
  pages = {7740--7754},
  doi = {10.18653/v1/2020.emnlp-main.623},
  url = {https://aclanthology.org/2020.emnlp-main.623/}
}

@misc{kweon2024kormedmcqa,
  author = {Kweon, Sunjun and Choi, Byungjin and Chu, Gyouk and Song, Junyeong and Hyeon, Daeun and Gan, Sujin and Kim, Jueon and Kim, Minkyu and Park, Rae Woong and Choi, Edward},
  title = {{KorMedMCQA}: multi-choice question answering benchmark for {Korean} healthcare professional licensing examinations},
  year = {2024},
  doi = {10.48550/arXiv.2403.01469},
  eprint = {2403.01469},
  archiveprefix = {arXiv},
  note = {arXiv:2403.01469}
}

@inproceedings{labrak2022frenchmedmcqa,
  author = {Labrak, Yanis and Bazoge, Adrien and Dufour, Richard and Daille, Béatrice and Gourraud, Pierre-Antoine and Morin, Emmanuel and Rouvier, Mickael},
  title = {{FrenchMedMCQA}: a {French} multiple-choice question answering dataset for medical domain},
  booktitle = {Proceedings of the 13th International Workshop on Health Text Mining and Information Analysis ({LOUHI})},
  year = {2022},
  pages = {41--46},
  doi = {10.18653/v1/2022.louhi-1.5}
}

@article{lau2018dataset,
  author = {Lau, Jason J and Gayen, Soumya and Ben Abacha, Asma and Demner-Fushman, Dina},
  title = {A dataset of clinically generated visual questions and answers about radiology images},
  journal = {Sci Data},
  year = {2018},
  volume = {5},
  pages = {180251},
  doi = {10.1038/sdata.2018.251},
  pmid = {30457565}
}

@article{lee2017curated,
  author = {Lee, Rebecca Sawyer and Gimenez, Francisco and Hoogi, Assaf and Miyake, Kanae Kawai and Gorovoy, Mia and Rubin, Daniel L},
  title = {A curated mammography data set for use in computer-aided detection and diagnosis research},
  journal = {Sci Data},
  year = {2017},
  volume = {4},
  pages = {170177},
  doi = {10.1038/sdata.2017.177},
  pmid = {29257132}
}

@article{lee2025lowdose,
  author = {Lee, Wonkyeong and Wagner, Fabian and Galdran, Adrian and Shi, Yongyi and Xia, Wenjun and Wang, Ge and Mou, Xuanqin and Ahamed, Md Atik and Imran, Abdullah Al Zubaer and Oh, Ji Eun and Kim, Kyungsang and Baek, Jong Tak and Lee, Dongheon and Hong, Boohwi and Tempelman, Philip and others},
  title = {Low-dose computed tomography perceptual image quality assessment},
  journal = {Med Image Anal},
  year = {2025},
  volume = {99},
  pages = {103343},
  doi = {10.1016/j.media.2024.103343},
  pmid = {39265362}
}

@article{leguellec2026parrot,
  author = {Le Guellec, Bastien and Adambounou, Kokou and Adams, Lisa C and Agripnidis, Thibault and Ahn, Sung Soo and Ait Chalal, Radhia and Akinci D'Antonoli, Tugba and Amouyel, Philippe and Andersson, Henrik and Bentegeac, Raphaël and Benzoni, Claudio and Blandino, Antonino Andrea and Busch, Felix and Can, Elif and Cau, Riccardo and others},
  title = {{PARROT}, an open multilingual radiology reports dataset},
  journal = {Eur J Radiol Artif Intell},
  year = {2026},
  volume = {5},
  pages = {100066},
  doi = {10.1016/j.ejrai.2025.100066}
}

@inproceedings{liu2021slake,
  author = {Liu, Bo and Zhan, Li-Ming and Xu, Li and Ma, Lin and Yang, Yan and Wu, Xiao-Ming},
  title = {{SLAKE}: a semantically-labeled knowledge-enhanced dataset for medical visual question answering},
  booktitle = {2021 IEEE 18th International Symposium on Biomedical Imaging (ISBI)},
  year = {2021},
  pages = {1650--1654},
  doi = {10.1109/ISBI48211.2021.9434010},
  note = {arXiv:2102.09542}
}

@article{liu2022deepdrid,
  author = {Liu, Ruhan and Wang, Xiangning and Wu, Qiang and Dai, Ling and Fang, Xi and Yan, Tao and Son, Jaemin and Tang, Shiqi and Li, Jiang and Gao, Zijian and Galdran, Adrian and Poorneshwaran, J M and Liu, Hao and Wang, Jie and Chen, Yerui and others},
  title = {{DeepDRiD}: diabetic retinopathy-grading and image quality estimation challenge},
  journal = {Patterns (N Y)},
  year = {2022},
  volume = {3},
  number = {6},
  pages = {100512},
  doi = {10.1016/j.patter.2022.100512},
  pmid = {35755875}
}

@inproceedings{liu2023benchmarking,
  author = {Liu, Junling and Zhou, Peilin and Hua, Yining and Chong, Dading and Tian, Zhongyu and Liu, Andrew and Wang, Helin and You, Chenyu and Guo, Zhenhua and Zhu, Lei and Li, Michael Lingzhi},
  title = {Benchmarking large language models on {CMExam} -- a comprehensive {Chinese} medical exam dataset},
  booktitle = {Advances in Neural Information Processing Systems (Datasets and Benchmarks Track)},
  year = {2023},
  volume = {36},
  pages = {52430--52452},
  doi = {10.48550/arXiv.2306.03030},
  url = {https://papers.nips.cc/paper_files/paper/2023/hash/a48ad12d588c597f4725a8b84af647b5-Abstract-Datasets_and_Benchmarks.html},
  note = {arXiv:2306.03030}
}

@inproceedings{nakamura2023ntcir17,
  author = {Nakamura, Yuta and Hanaoka, Shouhei and Yada, Shuntaro and Wakamiya, Shoko and Aramaki, Eiji},
  title = {{NTCIR-17 MedNLP-SC} radiology report subtask overview: dataset and solutions for automated lung cancer staging},
  booktitle = {Proceedings of the 17th {NTCIR} Conference on Evaluation of Information Access Technologies},
  year = {2023},
  pages = {145--151},
  doi = {10.20736/0002001328}
}

@article{narai2022movement,
  author = {N{\'a}rai, {\'A}d{\'a}m and Hermann, Petra and Auer, Tibor and Kemenczky, P{\'e}ter and Szalma, J{\'a}nos and Homolya, Istv{\'a}n and Somogyi, Eszter and Vakli, P{\'a}l and Weiss, B{\'e}la and Vidny{\'a}nszky, Zolt{\'a}n},
  title = {Movement-related artefacts ({MR-ART}) dataset of matched motion-corrupted and clean structural {MRI} brain scans},
  journal = {Sci Data},
  year = {2022},
  volume = {9},
  number = {1},
  pages = {630},
  doi = {10.1038/s41597-022-01694-8},
  pmid = {36253426}
}

@inproceedings{pal2022medmcqa,
  author = {Pal, Ankit and Umapathi, Logesh Kumar and Sankarasubbu, Malaikannan},
  title = {{MedMCQA}: a large-scale multi-subject multi-choice dataset for medical domain question answering},
  booktitle = {Proceedings of the Conference on Health, Inference, and Learning},
  series = {Proceedings of Machine Learning Research},
  year = {2022},
  volume = {174},
  pages = {248--260},
  doi = {10.48550/arXiv.2203.14371},
  url = {https://proceedings.mlr.press/v174/pal22a.html},
  note = {arXiv:2203.14371}
}

@article{porwal2018indian,
  author = {Porwal, Prasanna and Pachade, Samiksha and Kamble, Ravi and Kokare, Manesh and Deshmukh, Girish and Sahasrabuddhe, Vivek and Meriaudeau, Fabrice},
  title = {Indian {Diabetic} {Retinopathy} {Image} {Dataset} ({IDRiD}): a database for diabetic retinopathy screening research},
  journal = {Data},
  year = {2018},
  volume = {3},
  number = {3},
  pages = {25},
  doi = {10.3390/data3030025}
}

@article{richards2026rsna,
  author = {Richards, Tyler J and Flanders, Adam E and Colak, Errol and Prevedello, Luciano M and Ball, Robyn L and Kitamura, Felipe and Mongan, John and Vazirabad, Maryam and Lin, Hui-Ming and Kendell, Anne and Kanthawang, Thanat and Angkurawaranon, Salita and Altinmakas, Emre and Dogan, Hakan and Kuriki, Paulo Eduardo de Aguiar and others},
  title = {The {RSNA} {Lumbar} {Degenerative} {Imaging} {Spine} {Classification} ({LumbarDISC}) dataset},
  journal = {Radiol Artif Intell},
  year = {2026},
  volume = {8},
  number = {2},
  pages = {e250480},
  doi = {10.1148/ryai.250480},
  pmid = {41532835}
}

@inproceedings{roberts2022overview,
  author = {Roberts, Kirk and Demner-Fushman, Dina and Voorhees, Ellen M and Bedrick, Steven and Hersh, William R},
  title = {Overview of the {TREC} 2022 {Clinical Trials} track},
  booktitle = {The Thirty-First Text REtrieval Conference Proceedings (TREC 2022)},
  series = {NIST Special Publication 500-338},
  year = {2022},
  doi = {10.6028/NIST.SP.500-338.trials-overview},
  url = {https://trec.nist.gov/pubs/trec31/papers/Overview_trials.pdf}
}

@article{rotemberg2021patient,
  author = {Rotemberg, Veronica and Kurtansky, Nicholas and Betz-Stablein, Brigid and Caffery, Liam and Chousakos, Emmanouil and Codella, Noel and Combalia, Marc and Dusza, Stephen and Guitera, Pascale and Gutman, David and Halpern, Allan and Helba, Brian and Kittler, Harald and Kose, Kivanc and Langer, Steve and others},
  title = {A patient-centric dataset of images and metadata for identifying melanomas using clinical context},
  journal = {Sci Data},
  year = {2021},
  volume = {8},
  number = {1},
  pages = {34},
  doi = {10.1038/s41597-021-00815-z},
  pmid = {33510154}
}

@article{ruckert2024rocov2,
  author = {R{\"u}ckert, Johannes and Bloch, Louise and Br{\"u}ngel, Raphael and Idrissi-Yaghir, Ahmad and Sch{\"a}fer, Henning and Schmidt, Cynthia S and Koitka, Sven and Pelka, Obioma and Ben Abacha, Asma and Seco de Herrera, Alba G and M{\"u}ller, Henning and Horn, Peter A and Nensa, Felix and Friedrich, Christoph M},
  title = {{ROCOv2}: {Radiology Objects in COntext} version 2, an updated multimodal image dataset},
  journal = {Sci Data},
  year = {2024},
  volume = {11},
  number = {1},
  pages = {688},
  doi = {10.1038/s41597-024-03496-6},
  pmid = {38926396}
}

@misc{sawyerlee2016curated,
  author = {Sawyer-Lee, Rebecca and Gimenez, Francisco and Hoogi, Assaf and Rubin, Daniel},
  title = {Curated {Breast} {Imaging} {Subset} of {Digital} {Database} for {Screening} {Mammography} ({CBIS-DDSM})},
  year = {2016},
  publisher = {The Cancer Imaging Archive},
  version = {1},
  doi = {10.7937/K9/TCIA.2016.7O02S9CY},
  note = {Data set}
}

@inproceedings{segurabedmar2013semeval,
  author = {Segura-Bedmar, Isabel and Mart{\'i}nez, Paloma and Herrero-Zazo, Mar{\'i}a},
  title = {{SemEval-2013} Task 9: extraction of drug-drug interactions from biomedical texts ({DDIExtraction} 2013)},
  booktitle = {Second Joint Conference on Lexical and Computational Semantics (*SEM), Volume 2: Proceedings of the Seventh International Workshop on Semantic Evaluation (SemEval 2013)},
  year = {2013},
  pages = {341--350},
  url = {https://aclanthology.org/S13-2056/}
}

@article{sekuboyina2021verse,
  author = {Sekuboyina, Anjany and Husseini, Malek E and Bayat, Amirhossein and L{\"o}ffler, Maximilian and Liebl, Hans and Li, Hongwei and Tetteh, Giles and Kuka{\v{c}}ka, Jan and Payer, Christian and {\v{S}}tern, Darko and Urschler, Martin and Chen, Maodong and Cheng, Dalong and Lessmann, Nikolas and Hu, Yujin and others},
  title = {{VerSe}: a vertebrae labelling and segmentation benchmark for multi-detector {CT} images},
  journal = {Med Image Anal},
  year = {2021},
  volume = {73},
  pages = {102166},
  doi = {10.1016/j.media.2021.102166},
  pmid = {34340104}
}

@article{shoham2024medconceptsqa,
  author = {Shoham, Ofir Ben and Rappoport, Nadav},
  title = {{MedConceptsQA}: open source medical concepts {QA} benchmark},
  journal = {Comput Biol Med},
  year = {2024},
  volume = {182},
  pages = {109089},
  doi = {10.1016/j.compbiomed.2024.109089},
  pmid = {39276611}
}

@article{tang2021clip,
  author = {Tang, Jennifer S N and Seah, Jarrel C Y and Zia, Adil and Gajera, Jay and Schlegel, Richard N and Wong, Aaron J N and Gai, Dayu and Su, Shu and Bose, Tony and Kok, Marcus L and Jarema, Alex and Harisis, George N and Cheng, Chris-Tin and Kavnoudias, Helen and Wang, Wayland and others},
  title = {{CLiP}, catheter and line position dataset},
  journal = {Sci Data},
  year = {2021},
  volume = {8},
  number = {1},
  pages = {285},
  doi = {10.1038/s41597-021-01066-8},
  pmid = {34711836}
}

@misc{unifesp2022xray,
  author = {{Universidade Federal de S{\~a}o Paulo (UNIFESP)}},
  title = {{UNIFESP X-ray Body Part Classifier Competition}},
  year = {2022},
  url = {https://www.kaggle.com/datasets/felipekitamura/unifesp-xray-bodypart-classification},
  note = {Accessed October 2026}
}

@article{vazqueznoguera2025mammography,
  author = {V{\'a}zquez Noguera, Jos{\'e} Luis and Torres-Hurtado, Alejandro and G{\'o}mez-Adorno, Helena and Mello-Rom{\'a}n, Julio C{\'e}sar and Fleitas-Alvarez, Enrique Javier and Espinola Schulze, Federico Fernando and Garcia-Torres, Miguel and M{\'e}ndez Gaona, Carlos Domingo and Gardel Sotomayor, Pedro Esteban and V{\'a}zquez Noguera, Silvia and Zaracho Amarilla, Norma Elizabeth and Gamarra Esquivel, Oxades Wilfrido},
  title = {Mammography reporting dataset with {BI-RADS} system for natural language processing applications: addressing public data gaps in {Spanish}},
  journal = {Data Brief},
  year = {2025},
  volume = {61},
  pages = {111761},
  doi = {10.1016/j.dib.2025.111761},
  pmid = {40612472}
}

@inproceedings{vilares2019headqa,
  author = {Vilares, David and Gómez-Rodríguez, Carlos},
  title = {{HEAD-QA}: a healthcare dataset for complex reasoning},
  booktitle = {Proceedings of the 57th Annual Meeting of the Association for Computational Linguistics},
  year = {2019},
  pages = {960--966},
  doi = {10.18653/v1/P19-1092}
}

@inproceedings{wang2024mmlupro,
  author = {Wang, Yubo and Ma, Xueguang and Zhang, Ge and Ni, Yuansheng and Chandra, Abhranil and Guo, Shiguang and Ren, Weiming and Arulraj, Aaran and He, Xuan and Jiang, Ziyan and Li, Tianle and Ku, Max and Wang, Kai and Zhuang, Alex and Fan, Rongqi and others},
  title = {{MMLU-Pro}: a more robust and challenging multi-task language understanding benchmark},
  booktitle = {Advances in Neural Information Processing Systems (Datasets and Benchmarks Track)},
  year = {2024},
  volume = {37},
  pages = {95266--95290},
  doi = {10.48550/arXiv.2406.01574},
  url = {https://papers.nips.cc/paper_files/paper/2024/hash/ad236edc564f3e3156e1b2feafb99a24-Abstract-Datasets_and_Benchmarks_Track.html},
  note = {arXiv:2406.01574}
}

@article{yamagishi2025development,
  author = {Yamagishi, Yosuke and Nakamura, Yuta and Kikuchi, Tomohiro and Sonoda, Yuki and Hirakawa, Hiroshi and Kano, Shintaro and Nakamura, Satoshi and Hanaoka, Shouhei and Yoshikawa, Takeharu and Abe, Osamu},
  title = {Development of a large-scale dataset of chest computed tomography reports in {Japanese} and a high-performance finding classification model: dataset development and validation study},
  journal = {JMIR Med Inform},
  year = {2025},
  volume = {13},
  pages = {e71137},
  doi = {10.2196/71137},
  pmid = {40874833}
}

@article{zhang2024development,
  author = {Zhang, Xiaoman and Wu, Chaoyi and Zhao, Ziheng and Lin, Weixiong and Zhang, Ya and Wang, Yanfeng and Xie, Weidi},
  title = {Development of a large-scale medical visual question-answering dataset},
  journal = {Commun Med (Lond)},
  year = {2024},
  volume = {4},
  number = {1},
  pages = {277},
  doi = {10.1038/s43856-024-00709-2},
  pmid = {39709495}
}

@inproceedings{zhu2026diagnosisarena,
  author = {Zhu, Yakun and Huang, Zhongzhen and Mu, Linjie and Huang, Yutong and Nie, Wei and Liu, Jiaji and Zhang, Shaoting and Liu, Pengfei and Zhang, Xiaofan},
  title = {{DiagnosisArena}: benchmarking diagnostic reasoning for large language models},
  booktitle = {Findings of the Association for Computational Linguistics: ACL 2026},
  year = {2026},
  pages = {3074--3098},
  doi = {10.18653/v1/2026.findings-acl.151},
  url = {https://aclanthology.org/2026.findings-acl.151/},
  note = {arXiv:2505.14107}
}

@misc{zhang2026generalpurposedecisionmodelclinical,
      title={A general-purpose System One decision model for clinical records}, 
      author={Andrew Zhang and Trinh T. L. Vuong and Sophia J. Wagner and Dongmin Bang and Tong Ding and Yuchang Su and Mary Cipperman and Christina S. Chen and Judith Lefkes and Ramin Kahidi and Ming Y. Lu and Caiwei Tian and Tianhao Luo and Jingnan Ma and Rowland Pettit and Long Phi Le and Faisal Mahmood},
      year={2026},
      eprint={2610.27099},
      archivePrefix={arXiv},
      primaryClass={cs.AI},
      url={https://arxiv.org/abs/2610.27099}, 
}

@misc{qwen38,
    title = {{Qwen3.8-Max}: A New Bar for Coding and Cowork},
    url = {https://qwen.ai/blog?id=qwen3.8},
    author = {{Qwen Team}},
    month = {August},
    year = {2026}
}

@misc{qwen3.5,
    title  = {{Qwen3.5}: Towards Native Multimodal Agents},
    author = {{Qwen Team}},
    month  = {February},
    year   = {2026},
    url    = {https://qwen.ai/blog?id=qwen3.5}
}

@misc{cedars,
  author       = {{Akinci D'Antonoli}, Tugba and Doo, Florence},
  title        = {{CEDARS}: Carbon, Energy Diagnostics and Reporting for Sustainability},
  year         = {2026},
  version      = {1.0.0},
  howpublished = {Software},
  url          = {https://github.com/takinci/cedars},
  note         = {Apache-2.0. Accessed October 2026}
}

@misc{wei2023chainofthoughtpromptingelicitsreasoning,
      title={Chain-of-Thought Prompting Elicits Reasoning in Large Language Models}, 
      author={Jason Wei and Xuezhi Wang and Dale Schuurmans and Maarten Bosma and Brian Ichter and Fei Xia and Ed Chi and Quoc Le and Denny Zhou},
      year={2023},
      eprint={2201.11903},
      archivePrefix={arXiv},
      primaryClass={cs.CL},
      url={https://arxiv.org/abs/2201.11903}, 
}

\clearpage
\onecolumn
\appendix
\renewcommand{\thetable}{S\arabic{table}}
\renewcommand{\thefigure}{S\arabic{figure}}
\renewcommand{\thesection}{S\arabic{section}}
\setcounter{section}{0}\setcounter{figure}{0}\setcounter{table}{0}
\titleformat{\section}{\needspace{4\baselineskip}\sffamily\bfseries\normalsize}{\thesection}{1em}{}
\titleformat{\subsection}{\normalfont\bfseries}{\thesubsection}{1em}{}
\titlespacing*{\subsection}{0pt}{1.4ex plus .3ex minus .2ex}{0.6ex}
\setlength{\tabcolsep}{3pt}
\begin{center}
{\fontsize{18}{22}\selectfont Supplementary Appendix\par}
\vspace{0.4em}
{\large The Value of System~1 Models in Medicine\par}
\end{center}
\vspace{1em}
\captionsetup{list=yes}
\tableofcontents
\listoffigures
\listoftables
\clearpage

\section{Task Selection}
The taxonomy splits radiology into five layers (acquisition and metadata, image quality, anatomy, findings, and
examination or patient level). Table~\ref{tab:inventory} lists the evaluated tasks.

\section{Requests}
The state contained the case text as released, such as the question stem, the report sections or the triage fields,
or a one-sentence description of the input together with the image. Each option of a choice question had a short
description, and each level of an ordered score was described in words. Images larger than 4~MiB were downsampled by
half until they were below 4~MiB. For Jev~1.14, images with a side below 280 or above 1,568 pixels were rescaled. The
four 2D imaging tasks with higher-resolution images for Jev~1.14 were melanoma in ISIC 2020, paediatric pneumonia,
retinal OCT and diabetic retinopathy grade in RetinaMNIST. On OpenRouter, Clef and Clef-flash were routed only to the
Cloudflare provider, because another provider ignored images. When a model refused a question, the question was
removed, the request was repeated with the remaining questions, and the refusal was logged. Volumetric slices were taken within the relevant body region, with the window named
in the text.

\section{Scoring}
For yes-or-no questions, the AUROC used the probability of the positive option. For other questions, the macro
average included only the options present in the reference. Balanced accuracy was the recall of the most probable
option, averaged over the reference classes. Probabilities were normalised to sum to one, because two models round
probabilities to two decimals. The expected calibration error used 10 equal-width bins of the top-label probability.
When a task shared its requests with another task (chest radiograph findings and view, head CT haemorrhage and critical
findings), each question was counted in one task only.

\section{Generative Comparison}
The generative models replied in JSON. Qwen3.8-27B answered the first 50 of the 100 cases per task. Reasoning tokens
were counted and billed.

\section{Reference Results}
Table~\ref{tab:references} lists every published result with its type.

\section{Supplementary Results}

\subsection{All Tasks and Models}
\begin{figure}[p]\centering
\includegraphics[width=\textwidth]{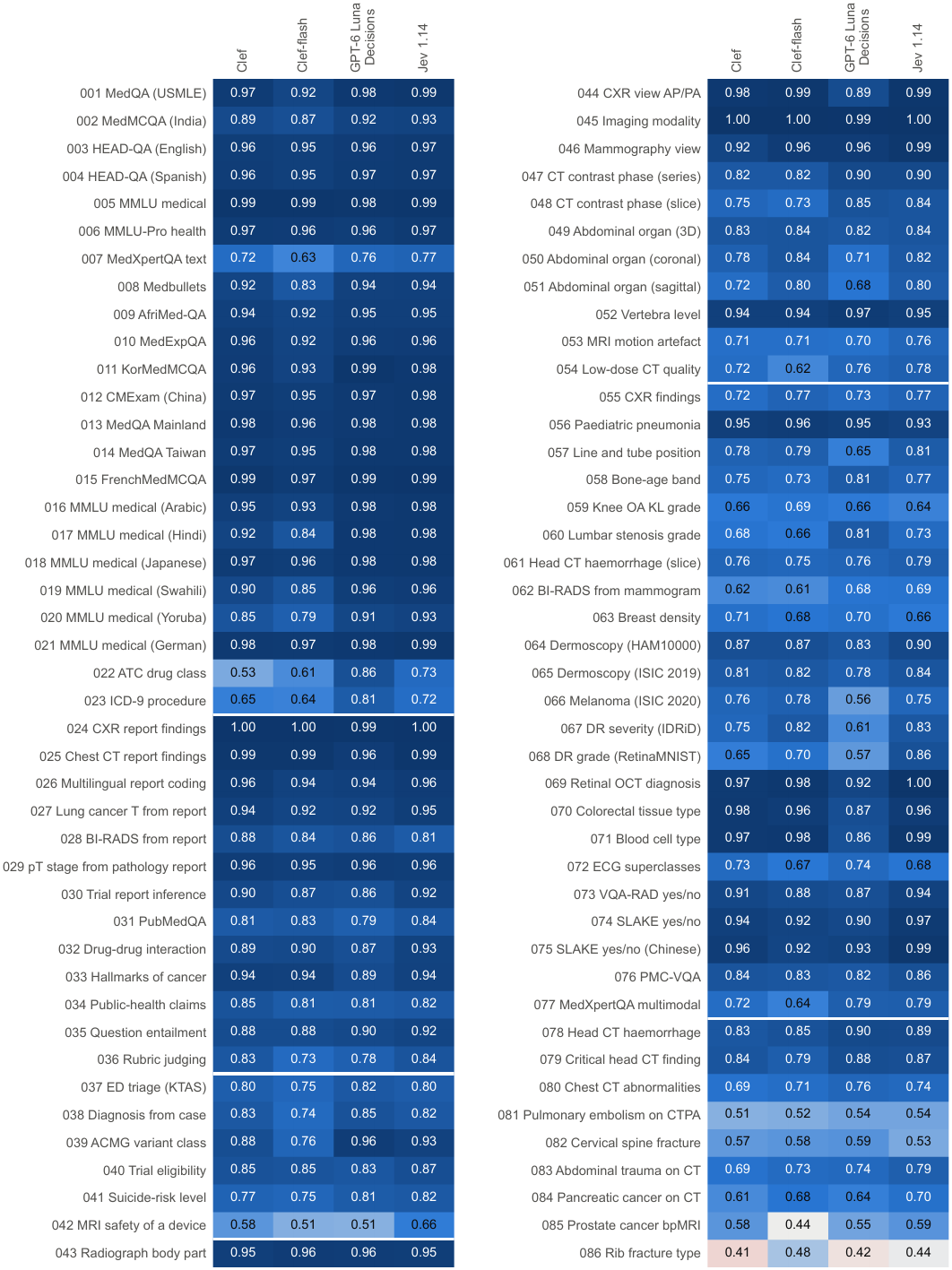}
\caption[AUROC of every task and model]{AUROC of every task (rows, numbered as in Table~\ref{tab:inventory}) and model (columns). Blue shades mark
values above chance (0.5) and red shades values below chance.}\label{fig:heatmap}
\end{figure}

Figure~\ref{fig:heatmap} shows the AUROC of every task for every model. Table~\ref{tab:per_task} gives the AUROC with
95\% CIs, and Table~\ref{tab:tiers} the random-effects means per tier.

\subsection{System~1 Models and Their Backbones}
\begin{figure}[p]\centering
\includegraphics[width=\textwidth]{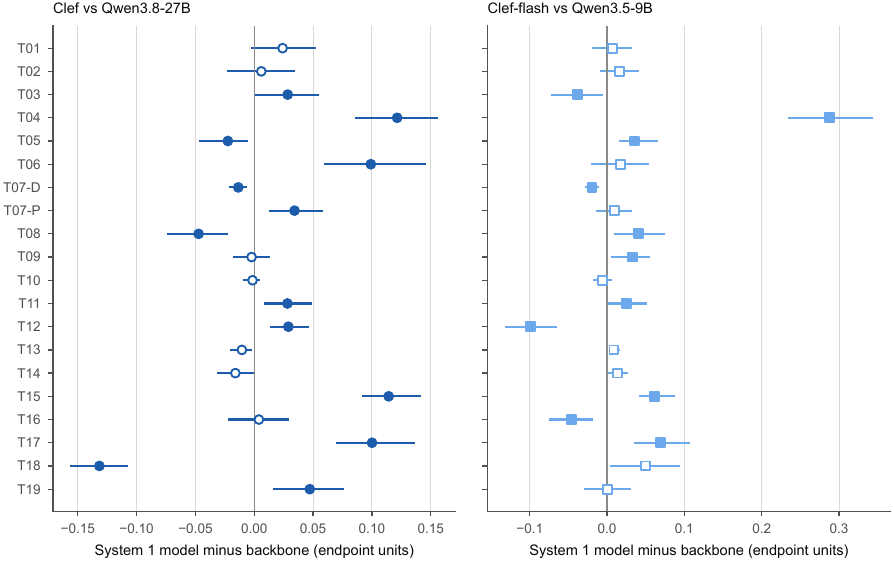}
\caption[System~1 models against their backbones]{Primary endpoint of each System~1 model minus that of its backbone on 20 tasks, with
95\% bias-corrected and accelerated bootstrap CIs. Filled markers indicate Holm-adjusted P$<$0.05 within the task.}\label{fig:backbone}
\end{figure}

On 20 tasks, Clef and Clef-flash were also compared with their backbones, Qwen3.8-27B and Qwen3.5-9B, with the
backbone read out over the same options (Figure~\ref{fig:backbone}). Each task had one primary endpoint. The
sign test across tasks was not significant for Clef-flash (better on 14 tasks and worse on 5, Holm-adjusted P=0.13)
or Clef (12 and 7, Holm-adjusted P=0.36). Five of the 16 significant advantages of a System~1 model disappeared or
reversed after the backbone readout was corrected for its option prior.

\subsection{Pooling of Slices}\label{sec:pooling}
\begin{figure}[p]\centering
\includegraphics[width=\textwidth]{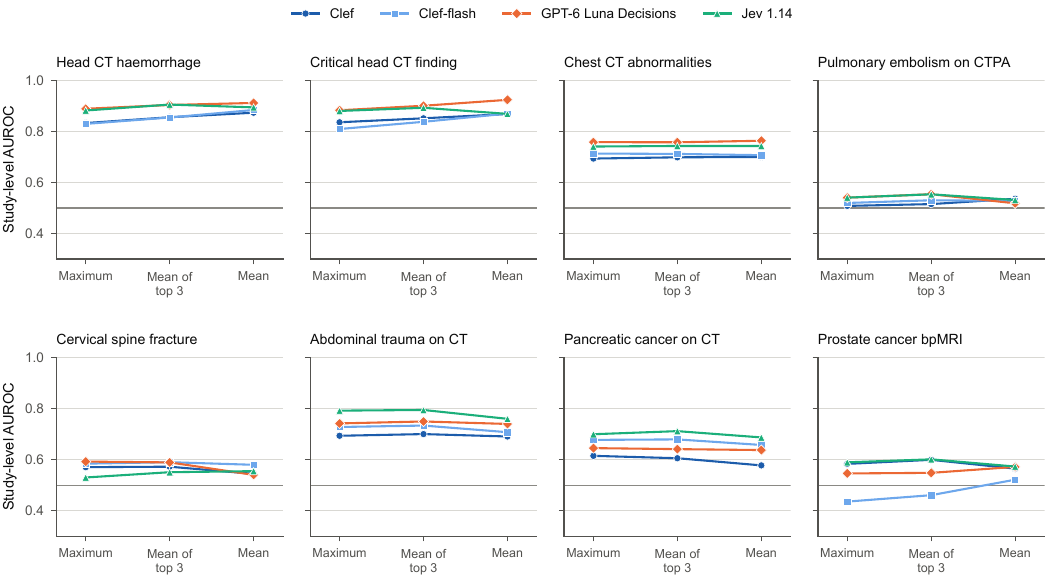}
\caption[Pooling of slices]{Study-level AUROC of volumetric tasks with yes-or-no questions, averaged over questions, when the slice
probabilities of a study were combined by their maximum (primary rule), the mean of the three highest, or the mean
over all slices. Each panel shows one task and each line one model.}\label{fig:pooling}
\end{figure}

For volumetric tasks, the mean study-level AUROC of yes-or-no questions over tasks and models was 0.68 with the maximum
over slices, 0.69 with the mean of the three highest slices and 0.69 with the mean over slices
(Figure~\ref{fig:pooling}). Pulmonary embolism and cervical spine fracture stayed close to chance under every rule.

\subsection{Generative Comparison}
Table~\ref{tab:generative} lists the balanced accuracy and cost of each System~1 model and its generative model per
task.

\clearpage

\begin{landscape}
\setlength{\LTcapwidth}{\textwidth}
{\footnotesize
\begin{longtable}{lP{9cm}P{4.2cm}llllrr}
\caption[Evaluated tasks]{Tasks. Tasks 004, 016, 017, 018, 019, 020, 021, 075 are language versions of another task. MCQ denotes multiple-choice questions, VQA visual question answering, and ED emergency department.}\label{tab:inventory}\\
\toprule
No. & Task & Dataset & Tier & Input & Language & Cases & Questions \\
\midrule
\endfirsthead
\toprule
No. & Task & Dataset & Tier & Input & Language & Cases & Questions \\
\midrule
\endhead
001 & USMLE-style MCQ (MedQA, 4 options) & MedQA\citep{jin2021disease} & Medical knowledge & text & English & 1,273 & 1 \\
\rowcolor{rowshade}002 & Indian medical entrance exam MCQ (MedMCQA) & MedMCQA\citep{pal2022medmcqa} & Medical knowledge & text & English & 4,183 & 1 \\
003 & Spanish healthcare specialisation exam MCQ (HEAD-QA) & HEAD-QA\citep{vilares2019headqa} & Medical knowledge & text & English & 2,675 & 1 \\
\rowcolor{rowshade}004 & Spanish healthcare specialisation exam MCQ, Spanish original (HEAD-QA) & HEAD-QA\citep{vilares2019headqa} & Medical knowledge & text & Spanish & 2,675 & 1 \\
005 & Medical subsets of MMLU & MMLU\citep{hendrycks2021measuring} & Medical knowledge & text & English & 1,089 & 1 \\
\rowcolor{rowshade}006 & Health questions of MMLU-Pro & MMLU-Pro\citep{wang2024mmlupro} & Medical knowledge & text & English & 818 & 1 \\
007 & Expert-level medical MCQ, text (MedXpertQA Text) & MedXpertQA\citep{zuo2025medxpertqa} & Medical knowledge & text & English & 2,450 & 1 \\
\rowcolor{rowshade}008 & Hard clinical case MCQ (Medbullets op5) & Medbullets op5\citep{chen2025benchmarking} & Medical knowledge & text & English & 308 & 1 \\
009 & Pan-African medical MCQ (AfriMed-QA v2) & AfriMed-QA v2\citep{nimo2025afrimed} & Medical knowledge & text & English & 3,600 & 1 \\
\rowcolor{rowshade}010 & Multilingual clinical-case MCQ (MedExpQA) & MedExpQA\citep{alonso2024medexpqa} & Medical knowledge & text & Several & 500 & 1 \\
011 & Korean healthcare licensing exam MCQ (KorMedMCQA) & KorMedMCQA\citep{kweon2024kormedmcqa} & Medical knowledge & text & Korean & 3,009 & 1 \\
\rowcolor{rowshade}012 & Chinese national medical licensing exam MCQ (CMExam) & CMExam\citep{liu2023benchmarking} & Medical knowledge & text & Chinese & 6,607 & 1 \\
013 & Chinese medical licensing exam MCQ (MedQA Mainland) & MedQA\citep{jin2021disease} & Medical knowledge & text & Chinese & 3,426 & 1 \\
\rowcolor{rowshade}014 & Taiwanese medical licensing exam MCQ (MedQA Taiwan) & MedQA\citep{jin2021disease} & Medical knowledge & text & Chinese & 1,413 & 1 \\
015 & French pharmacy exam MCQ (FrenchMedMCQA) & FrenchMedMCQA\citep{labrak2022frenchmedmcqa} & Medical knowledge & text & French & 321 & 1 \\
\rowcolor{rowshade}016 & Medical subsets of MMLU in Arabic (Global-MMLU) & Global-MMLU\citep{singh2025global} & Medical knowledge & text & Arabic & 1,089 & 1 \\
017 & Medical subsets of MMLU in Hindi (Global-MMLU) & Global-MMLU\citep{singh2025global} & Medical knowledge & text & Hindi & 1,089 & 1 \\
\rowcolor{rowshade}018 & Medical subsets of MMLU in Japanese (Global-MMLU) & Global-MMLU\citep{singh2025global} & Medical knowledge & text & Japanese & 1,089 & 1 \\
019 & Medical subsets of MMLU in Swahili (Global-MMLU) & Global-MMLU\citep{singh2025global} & Medical knowledge & text & Swahili & 1,089 & 1 \\
\rowcolor{rowshade}020 & Medical subsets of MMLU in Yoruba (Global-MMLU) & Global-MMLU\citep{singh2025global} & Medical knowledge & text & Yoruba & 1,089 & 1 \\
021 & Medical subsets of MMLU in German (Global-MMLU) & Global-MMLU\citep{singh2025global} & Medical knowledge & text & German & 1,089 & 1 \\
\rowcolor{rowshade}022 & Drug class from ATC code (MedConceptsQA) & MedConceptsQA\citep{shoham2024medconceptsqa} & Medical knowledge & text & English & 18,806 & 1 \\
023 & Procedure description from ICD-9-CM code (MedConceptsQA) & MedConceptsQA\citep{shoham2024medconceptsqa} & Medical knowledge & text & English & 13,766 & 1 \\
\rowcolor{rowshade}024 & Chest radiograph report findings (OpenI / IU X-ray) & OpenI / IU X-ray\citep{demnerfushman2016preparing} & Text understanding & text & English & 3,832 & 13 \\
025 & Chest CT findings from Japanese reports (CT-RATE-JPN) & CT-RATE-JPN\citep{yamagishi2025development,hamamci2026generalist} & Text understanding & text & Japanese & 150 & 18 \\
\rowcolor{rowshade}026 & Modality and ICD-10 chapters of multilingual reports (PARROT) & PARROT\citep{leguellec2026parrot} & Text understanding & text & Several & 2,738 & 13 \\
027 & Lung cancer T category from radiology reports (NTCIR-17 RR-TNM) & NTCIR-17 RR-TNM\citep{nakamura2023ntcir17} & Text understanding & text & Japanese & 243 & 1 \\
\rowcolor{rowshade}028 & BI-RADS category from breast imaging reports (Paraguay corpus) & Paraguay reports\citep{vazqueznoguera2025mammography} & Text understanding & text & English & 1,000 & 1 \\
029 & Pathologic T stage from a pathology report (TCGA) & TCGA reports\citep{kefeli2024tcga,kefeli2024generalizable} & Text understanding & text & English & 1,000 & 1 \\
\rowcolor{rowshade}030 & Clinical-trial report inference (NLI4CT, SemEval-2024 Task 2) & NLI4CT\citep{jullien2024semeval} & Text understanding & text & English & 500 & 1 \\
031 & Biomedical research QA (PubMedQA PQA-L) & PubMedQA PQA-L\citep{jin2019pubmedqa} & Text understanding & text & English & 500 & 1 \\
\rowcolor{rowshade}032 & Drug-drug interaction type (DDI-2013) & DDI-2013\citep{herrerozazo2013ddi,segurabedmar2013semeval} & Text understanding & text & English & 1,000 & 1 \\
033 & Hallmarks of cancer in abstracts (HoC) & HoC\citep{baker2016automatic} & Text understanding & text & English & 371 & 10 \\
\rowcolor{rowshade}034 & Public-health claim veracity (PUBHEALTH) & PUBHEALTH\citep{kotonya2020explainable} & Text understanding & text & English & 1,233 & 1 \\
035 & Consumer health question entailment (MEDIQA-RQE 2019) & MEDIQA-RQE 2019\citep{benabacha2019overview} & Text understanding & text & English & 230 & 1 \\
\rowcolor{rowshade}036 & Rubric criterion judging (HealthBench meta-evaluation) & HealthBench\citep{arora2025healthbench} & Text understanding & text & English & 1,000 & 1 \\
037 & ED triage acuity (KTAS) & KTAS\citep{moon2019triage} & Clinical judgement from text & text & English & 1,267 & 1 \\
\rowcolor{rowshade}038 & Diagnosis selection from a published case (DiagnosisArena) & DiagnosisArena\citep{zhu2026diagnosisarena} & Clinical judgement from text & text & English & 915 & 1 \\
039 & ACMG/AMP five-tier variant classification (ClinGen) & ClinGen\citep{rehm2015clingen} & Clinical judgement from text & text & English & 1,000 & 1 \\
\rowcolor{rowshade}040 & Patient-trial eligibility (TREC Clinical Trials 2021/2022) & TREC CT 2021/22\citep{roberts2022overview} & Clinical judgement from text & text & English & 500 & 1 \\
041 & Suicide-risk severity of a Reddit user (Reddit C-SSRS) & Reddit C-SSRS\citep{gaur2019knowledge} & Clinical judgement from text & text & English & 500 & 1 \\
\rowcolor{rowshade}042 & MRI safety status of a device (FDA AccessGUDID) & FDA AccessGUDID\citep{fda2026accessgudid} & Clinical judgement from text & text & English & 1,000 & 1 \\
043 & Radiograph body part & UNIFESP body part\citep{unifesp2022xray} & Image perception & 2D & English & 593 & 1 \\
\rowcolor{rowshade}044 & Chest radiograph view AP vs PA (NIH ChestX-ray14) & NIH ChestX-ray14\citep{wang2017chestx} & Image perception & 2D & English & 716 & 1 \\
045 & Imaging modality (ROCOv2) & ROCOv2\citep{ruckert2024rocov2} & Image perception & 2D & English & 794 & 1 \\
\rowcolor{rowshade}046 & Mammography view CC vs MLO (CBIS-DDSM) & CBIS-DDSM\citep{lee2017curated,sawyerlee2016curated} & Image perception & 2D & English & 645 & 1 \\
047 & CT contrast phase (series level, WAW-TACE) & WAW-TACE\citep{bartnik2024waw} & Image perception & 3D & English & 160 & 1 \\
\rowcolor{rowshade}048 & Contrast status and phase (single slice, WAW-TACE) & WAW-TACE\citep{bartnik2024waw} & Image perception & 2D & English & 960 & 2 \\
049 & Abdominal organ on CT, 3D (OrganMNIST3D) & OrganMNIST3D\citep{yang2023medmnist,bilic2023liver} & Image perception & 2D multi-view & English & 610 & 1 \\
\rowcolor{rowshade}050 & Abdominal organ on coronal CT (OrganCMNIST) & OrganCMNIST\citep{yang2023medmnist,bilic2023liver} & Image perception & 2D & English & 8,216 & 1 \\
051 & Abdominal organ on sagittal CT (OrganSMNIST) & OrganSMNIST\citep{yang2023medmnist,bilic2023liver} & Image perception & 2D & English & 8,827 & 1 \\
\rowcolor{rowshade}052 & Marked vertebra level on CT (VerSe) & VerSe\citep{sekuboyina2021verse} & Image perception & 2D multi-view & English & 219 & 2 \\
053 & MRI motion artefact severity (MR-ART) & MR-ART\citep{narai2022movement} & Image perception & 3D & English & 436 & 1 \\
\rowcolor{rowshade}054 & Diagnostic image quality of low-dose CT (LDCTIQAC 2023) & LDCTIQAC 2023\citep{lee2025lowdose} & Image perception & 2D & English & 1,000 & 1 \\
055 & Chest radiograph multi-label findings (NIH ChestX-ray14) & NIH ChestX-ray14\citep{wang2017chestx} & 2D image findings & 2D & English & 716 & 14 \\
\rowcolor{rowshade}056 & Pneumonia on paediatric chest radiographs (PneumoniaMNIST) & PneumoniaMNIST\citep{yang2023medmnist,kermany2018identifying} & 2D image findings & 2D & English & 624 & 1 \\
057 & Line and tube position on chest radiographs & RANZCR CLiP\citep{tang2021clip,wang2017chestx} & 2D image findings & 2D & English & 497 & 11 \\
\rowcolor{rowshade}058 & Paediatric bone-age band & RSNA bone age\citep{halabi2019rsna} & 2D image findings & 2D & English & 500 & 1 \\
059 & Knee osteoarthritis KL grade & OAI (Chen 2018)\citep{chen2019fully,chen2018knee} & 2D image findings & 2D & English & 1,000 & 1 \\
\rowcolor{rowshade}060 & Lumbar degenerative stenosis grade & RSNA lumbar 2024\citep{richards2026rsna} & 2D image findings & 2D multi-view & English & 1,000 & 5 \\
061 & Head CT haemorrhage and subtype (slice) & RSNA ICH 2019\citep{flanders2020construction} & 2D image findings & 2D & English & 1,000 & 6 \\
\rowcolor{rowshade}062 & BI-RADS assessment from mammograms (CBIS-DDSM) & CBIS-DDSM\citep{lee2017curated,sawyerlee2016curated} & 2D image findings & 2D multi-view & English & 704 & 1 \\
063 & Breast density category (CBIS-DDSM) & CBIS-DDSM\citep{lee2017curated,sawyerlee2016curated} & 2D image findings & 2D & English & 643 & 1 \\
\rowcolor{rowshade}064 & Dermoscopic lesion diagnosis (HAM10000 / ISIC 2018 Task 3) & HAM10000\citep{codella2019skin,tschandl2018ham10000} & 2D image findings & 2D & English & 515 & 1 \\
065 & Dermoscopic lesion diagnosis (ISIC 2019) & ISIC 2019\citep{tschandl2018ham10000,codella2018skin,hernandezperez2024bcn20000} & 2D image findings & 2D & English & 8,238 & 1 \\
\rowcolor{rowshade}066 & Melanoma vs other lesion (SIIM-ISIC 2020) & SIIM-ISIC 2020\citep{rotemberg2021patient} & 2D image findings & 2D & English & 33,126 & 1 \\
067 & Diabetic retinopathy severity (IDRiD) & IDRiD\citep{porwal2018indian} & 2D image findings & 2D & English & 516 & 1 \\
\rowcolor{rowshade}068 & Diabetic retinopathy grade (RetinaMNIST) & RetinaMNIST\citep{yang2023medmnist,liu2022deepdrid} & 2D image findings & 2D & English & 400 & 1 \\
069 & Retinal OCT diagnosis (OCTMNIST) & OCTMNIST\citep{yang2023medmnist,kermany2018identifying} & 2D image findings & 2D & English & 1,000 & 1 \\
\rowcolor{rowshade}070 & Colorectal H\&E tissue type (CRC-VAL-HE-7K) & CRC-VAL-HE-7K\citep{kather2019predicting} & 2D image findings & 2D & English & 900 & 1 \\
071 & Blood cell type (BloodMNIST) & BloodMNIST\citep{yang2023medmnist,acevedo2020dataset} & 2D image findings & 2D & English & 3,421 & 1 \\
\rowcolor{rowshade}072 & 12-lead ECG superclasses (PTB-XL, rendered as image) & PTB-XL\citep{wagner2020ptb} & 2D image findings & 2D & English & 2,198 & 5 \\
073 & Radiology VQA, closed yes/no questions (VQA-RAD) & VQA-RAD\citep{lau2018dataset} & 2D image findings & text+image & English & 251 & 1 \\
\rowcolor{rowshade}074 & Radiology VQA, closed yes/no questions (SLAKE) & SLAKE\citep{liu2021slake} & 2D image findings & text+image & English & 355 & 1 \\
075 & Radiology VQA in Chinese, closed yes/no questions (SLAKE) & SLAKE\citep{liu2021slake} & 2D image findings & text+image & Chinese & 347 & 1 \\
\rowcolor{rowshade}076 & Biomedical figure VQA (PMC-VQA) & PMC-VQA\citep{zhang2024development} & 2D image findings & text+image & English & 2,000 & 1 \\
077 & Expert-level medical MCQ with images (MedXpertQA MM) & MedXpertQA\citep{zuo2025medxpertqa} & 2D image findings & text+image & English & 2,000 & 1 \\
\rowcolor{rowshade}078 & Head CT haemorrhage and subtype (scan level, CQ500) & CQ500\citep{chilamkurthy2018deep} & Volumetric findings & 3D & English & 81 & 6 \\
079 & Critical head CT finding (CQ500) & CQ500\citep{chilamkurthy2018deep} & Volumetric findings & 3D & English & 81 & 4 \\
\rowcolor{rowshade}080 & Chest CT multi-abnormality (CT-RATE) & CT-RATE\citep{hamamci2026generalist} & Volumetric findings & 3D & English & 100 & 18 \\
081 & Pulmonary embolism on CTPA & RSNA PE 2020\citep{colak2021rsna} & Volumetric findings & 3D & English & 150 & 6 \\
\rowcolor{rowshade}082 & Cervical spine fracture on CT & RSNA C-spine 2022\citep{lin2023rsna} & Volumetric findings & 3D & English & 200 & 8 \\
083 & Traumatic abdominal organ injury on CT & RSNA trauma 2023\citep{rudie2024rsna} & Volumetric findings & 3D & English & 200 & 8 \\
\rowcolor{rowshade}084 & Pancreatic ductal adenocarcinoma on contrast-enhanced CT (PANORAMA) & PANORAMA\citep{alves2026artificial} & Volumetric findings & 3D & English & 300 & 1 \\
085 & Clinically significant prostate cancer on bpMRI (PI-CAI) & PI-CAI\citep{saha2024artificial} & Volumetric findings & 3D & English & 200 & 1 \\
\rowcolor{rowshade}086 & Rib fracture type (FractureMNIST3D, RibFrac) & FractureMNIST3D\citep{yang2023medmnist,jin2020deep} & Volumetric findings & 2D multi-view & English & 240 & 1 \\
\bottomrule
\end{longtable}
}

\clearpage
{\footnotesize
\begin{longtable}{llcccc}
\caption[AUROC per task and model]{AUROC per task and model with 95\% CI from 1,000 bootstrap resamples of cases. Refused or failed questions were scored as wrong with the label prevalence as probability.}\label{tab:per_task}\\
\toprule
No. & Task & Clef & Clef-flash & GPT-6 Luna Decisions & Jev 1.14 \\
\midrule
\endfirsthead
\toprule
No. & Task & Clef & Clef-flash & GPT-6 Luna Decisions & Jev 1.14 \\
\midrule
\endhead
001 & MedQA (USMLE) & 0.97 (0.96--0.97) & 0.92 (0.90--0.93) & 0.98 (0.97--0.98) & 0.99 (0.98--0.99) \\
\rowcolor{rowshade}002 & MedMCQA (India) & 0.89 (0.89--0.90) & 0.87 (0.86--0.88) & 0.92 (0.92--0.93) & 0.93 (0.93--0.94) \\
003 & HEAD-QA (English) & 0.96 (0.95--0.96) & 0.95 (0.94--0.95) & 0.96 (0.96--0.97) & 0.97 (0.96--0.97) \\
\rowcolor{rowshade}004 & HEAD-QA (Spanish) & 0.96 (0.95--0.97) & 0.95 (0.94--0.95) & 0.97 (0.96--0.97) & 0.97 (0.97--0.98) \\
005 & MMLU medical & 0.99 (0.98--0.99) & 0.99 (0.98--0.99) & 0.98 (0.98--0.99) & 0.99 (0.98--0.99) \\
\rowcolor{rowshade}006 & MMLU-Pro health & 0.97 (0.96--0.97) & 0.96 (0.95--0.97) & 0.96 (0.94--0.96) & 0.97 (0.96--0.98) \\
007 & MedXpertQA text & 0.72 (0.71--0.73) & 0.63 (0.62--0.64) & 0.76 (0.74--0.77) & 0.77 (0.76--0.78) \\
\rowcolor{rowshade}008 & Medbullets & 0.92 (0.90--0.94) & 0.83 (0.80--0.86) & 0.94 (0.92--0.96) & 0.94 (0.93--0.96) \\
009 & AfriMed-QA & 0.94 (0.94--0.95) & 0.92 (0.91--0.93) & 0.95 (0.95--0.96) & 0.95 (0.95--0.96) \\
\rowcolor{rowshade}010 & MedExpQA & 0.96 (0.95--0.97) & 0.92 (0.91--0.94) & 0.96 (0.94--0.97) & 0.96 (0.95--0.98) \\
011 & KorMedMCQA & 0.96 (0.96--0.97) & 0.93 (0.92--0.93) & 0.99 (0.99--0.99) & 0.98 (0.98--0.98) \\
\rowcolor{rowshade}012 & CMExam (China) & 0.97 (0.97--0.97) & 0.95 (0.95--0.96) & 0.97 (0.97--0.98) & 0.98 (0.98--0.99) \\
013 & MedQA Mainland & 0.98 (0.98--0.98) & 0.96 (0.96--0.97) & 0.98 (0.97--0.98) & 0.98 (0.98--0.99) \\
\rowcolor{rowshade}014 & MedQA Taiwan & 0.97 (0.96--0.98) & 0.95 (0.95--0.96) & 0.98 (0.98--0.99) & 0.98 (0.98--0.99) \\
015 & FrenchMedMCQA & 0.99 (0.98--0.99) & 0.97 (0.95--0.98) & 0.99 (0.98--1.00) & 0.99 (0.99--1.00) \\
\rowcolor{rowshade}016 & MMLU medical (Arabic) & 0.95 (0.94--0.96) & 0.93 (0.92--0.94) & 0.98 (0.97--0.98) & 0.98 (0.98--0.99) \\
017 & MMLU medical (Hindi) & 0.92 (0.91--0.93) & 0.84 (0.83--0.86) & 0.98 (0.97--0.98) & 0.98 (0.97--0.99) \\
\rowcolor{rowshade}018 & MMLU medical (Japanese) & 0.97 (0.96--0.97) & 0.96 (0.95--0.97) & 0.98 (0.97--0.98) & 0.98 (0.98--0.99) \\
019 & MMLU medical (Swahili) & 0.90 (0.89--0.92) & 0.85 (0.83--0.86) & 0.96 (0.95--0.96) & 0.96 (0.95--0.97) \\
\rowcolor{rowshade}020 & MMLU medical (Yoruba) & 0.85 (0.83--0.86) & 0.79 (0.77--0.81) & 0.91 (0.90--0.93) & 0.93 (0.92--0.94) \\
021 & MMLU medical (German) & 0.98 (0.97--0.98) & 0.97 (0.96--0.98) & 0.98 (0.97--0.99) & 0.99 (0.98--0.99) \\
\rowcolor{rowshade}022 & ATC drug class & 0.53 (0.52--0.54) & 0.61 (0.60--0.62) & 0.86 (0.86--0.87) & 0.73 (0.73--0.74) \\
023 & ICD-9 procedure & 0.65 (0.64--0.65) & 0.64 (0.63--0.65) & 0.81 (0.80--0.81) & 0.72 (0.71--0.72) \\
\rowcolor{rowshade}024 & CXR report findings & 1.00 (0.99--1.00) & 1.00 (0.99--1.00) & 0.99 (0.99--0.99) & 1.00 (1.00--1.00) \\
025 & Chest CT report findings & 0.99 (0.98--0.99) & 0.99 (0.98--0.99) & 0.96 (0.96--0.97) & 0.99 (0.99--0.99) \\
\rowcolor{rowshade}026 & Multilingual report coding & 0.96 (0.96--0.97) & 0.94 (0.93--0.94) & 0.94 (0.94--0.95) & 0.96 (0.96--0.97) \\
027 & Lung cancer T from report & 0.94 (0.92--0.96) & 0.92 (0.90--0.94) & 0.92 (0.89--0.95) & 0.95 (0.93--0.97) \\
\rowcolor{rowshade}028 & BI-RADS from report & 0.88 (0.87--0.90) & 0.84 (0.82--0.86) & 0.86 (0.82--0.89) & 0.81 (0.77--0.84) \\
029 & pT stage from pathology report & 0.96 (0.95--0.97) & 0.95 (0.94--0.96) & 0.96 (0.94--0.97) & 0.96 (0.95--0.97) \\
\rowcolor{rowshade}030 & Trial report inference & 0.90 (0.88--0.93) & 0.87 (0.84--0.90) & 0.86 (0.83--0.90) & 0.92 (0.90--0.95) \\
031 & PubMedQA & 0.81 (0.78--0.84) & 0.83 (0.79--0.86) & 0.79 (0.76--0.83) & 0.84 (0.81--0.87) \\
\rowcolor{rowshade}032 & Drug-drug interaction & 0.89 (0.88--0.91) & 0.90 (0.89--0.91) & 0.87 (0.85--0.89) & 0.93 (0.92--0.94) \\
033 & Hallmarks of cancer & 0.94 (0.93--0.95) & 0.94 (0.93--0.95) & 0.89 (0.87--0.91) & 0.94 (0.93--0.94) \\
\rowcolor{rowshade}034 & Public-health claims & 0.85 (0.83--0.87) & 0.81 (0.78--0.83) & 0.81 (0.79--0.84) & 0.82 (0.80--0.84) \\
035 & Question entailment & 0.88 (0.84--0.92) & 0.88 (0.84--0.92) & 0.90 (0.86--0.94) & 0.92 (0.88--0.95) \\
\rowcolor{rowshade}036 & Rubric judging & 0.83 (0.79--0.86) & 0.73 (0.69--0.77) & 0.78 (0.75--0.82) & 0.84 (0.81--0.88) \\
037 & ED triage (KTAS) & 0.80 (0.78--0.81) & 0.75 (0.73--0.77) & 0.82 (0.80--0.84) & 0.80 (0.78--0.82) \\
\rowcolor{rowshade}038 & Diagnosis from case & 0.83 (0.81--0.85) & 0.74 (0.71--0.76) & 0.85 (0.83--0.87) & 0.82 (0.80--0.84) \\
039 & ACMG variant class & 0.88 (0.87--0.89) & 0.76 (0.74--0.78) & 0.96 (0.95--0.96) & 0.93 (0.92--0.94) \\
\rowcolor{rowshade}040 & Trial eligibility & 0.85 (0.83--0.88) & 0.85 (0.82--0.88) & 0.83 (0.80--0.86) & 0.87 (0.84--0.90) \\
041 & Suicide-risk level & 0.77 (0.75--0.80) & 0.75 (0.73--0.78) & 0.81 (0.78--0.83) & 0.82 (0.79--0.84) \\
\rowcolor{rowshade}042 & MRI safety of a device & 0.58 (0.56--0.61) & 0.51 (0.49--0.54) & 0.51 (0.49--0.53) & 0.66 (0.64--0.68) \\
043 & Radiograph body part & 0.95 (0.94--0.96) & 0.96 (0.96--0.97) & 0.96 (0.95--0.96) & 0.95 (0.95--0.96) \\
\rowcolor{rowshade}044 & CXR view AP/PA & 0.98 (0.97--0.99) & 0.99 (0.98--0.99) & 0.89 (0.86--0.91) & 0.99 (0.99--0.99) \\
045 & Imaging modality & 1.00 (1.00--1.00) & 1.00 (1.00--1.00) & 0.99 (0.99--1.00) & 1.00 (1.00--1.00) \\
\rowcolor{rowshade}046 & Mammography view & 0.92 (0.90--0.94) & 0.96 (0.95--0.98) & 0.96 (0.95--0.98) & 0.99 (0.98--0.99) \\
047 & CT contrast phase (series) & 0.82 (0.79--0.85) & 0.82 (0.78--0.86) & 0.90 (0.87--0.93) & 0.90 (0.87--0.92) \\
\rowcolor{rowshade}048 & CT contrast phase (slice) & 0.75 (0.73--0.77) & 0.73 (0.70--0.75) & 0.85 (0.83--0.86) & 0.84 (0.82--0.86) \\
049 & Abdominal organ (3D) & 0.83 (0.82--0.85) & 0.84 (0.83--0.86) & 0.82 (0.81--0.84) & 0.84 (0.82--0.85) \\
\rowcolor{rowshade}050 & Abdominal organ (coronal) & 0.78 (0.78--0.79) & 0.84 (0.84--0.84) & 0.71 (0.70--0.71) & 0.82 (0.82--0.83) \\
051 & Abdominal organ (sagittal) & 0.72 (0.72--0.73) & 0.80 (0.80--0.81) & 0.68 (0.67--0.68) & 0.80 (0.80--0.81) \\
\rowcolor{rowshade}052 & Vertebra level & 0.94 (0.93--0.95) & 0.94 (0.93--0.95) & 0.97 (0.96--0.97) & 0.95 (0.94--0.96) \\
053 & MRI motion artefact & 0.71 (0.69--0.74) & 0.71 (0.68--0.73) & 0.70 (0.67--0.72) & 0.76 (0.73--0.78) \\
\rowcolor{rowshade}054 & Low-dose CT quality & 0.72 (0.70--0.74) & 0.62 (0.60--0.65) & 0.76 (0.74--0.78) & 0.78 (0.77--0.80) \\
055 & CXR findings & 0.72 (0.69--0.74) & 0.77 (0.75--0.79) & 0.73 (0.71--0.76) & 0.77 (0.75--0.79) \\
\rowcolor{rowshade}056 & Paediatric pneumonia & 0.95 (0.93--0.96) & 0.96 (0.94--0.97) & 0.95 (0.93--0.96) & 0.93 (0.91--0.95) \\
057 & Line and tube position & 0.78 (0.76--0.79) & 0.79 (0.78--0.80) & 0.65 (0.63--0.67) & 0.81 (0.79--0.82) \\
\rowcolor{rowshade}058 & Bone-age band & 0.75 (0.73--0.77) & 0.73 (0.71--0.75) & 0.81 (0.79--0.83) & 0.77 (0.75--0.78) \\
059 & Knee OA KL grade & 0.66 (0.64--0.68) & 0.69 (0.67--0.70) & 0.66 (0.63--0.69) & 0.64 (0.62--0.66) \\
\rowcolor{rowshade}060 & Lumbar stenosis grade & 0.68 (0.66--0.70) & 0.66 (0.65--0.68) & 0.81 (0.80--0.83) & 0.73 (0.71--0.75) \\
061 & Head CT haemorrhage (slice) & 0.76 (0.74--0.78) & 0.75 (0.73--0.77) & 0.76 (0.74--0.78) & 0.79 (0.77--0.80) \\
\rowcolor{rowshade}062 & BI-RADS from mammogram & 0.62 (0.59--0.65) & 0.61 (0.59--0.65) & 0.68 (0.65--0.70) & 0.69 (0.65--0.72) \\
063 & Breast density & 0.71 (0.68--0.73) & 0.68 (0.65--0.71) & 0.70 (0.67--0.73) & 0.66 (0.63--0.69) \\
\rowcolor{rowshade}064 & Dermoscopy (HAM10000) & 0.87 (0.85--0.89) & 0.87 (0.85--0.89) & 0.83 (0.81--0.85) & 0.90 (0.89--0.92) \\
065 & Dermoscopy (ISIC 2019) & 0.81 (0.81--0.82) & 0.82 (0.82--0.83) & 0.78 (0.77--0.79) & 0.84 (0.83--0.84) \\
\rowcolor{rowshade}066 & Melanoma (ISIC 2020) & 0.76 (0.74--0.78) & 0.78 (0.76--0.80) & 0.56 (0.53--0.59) & 0.75 (0.73--0.77) \\
067 & DR severity (IDRiD) & 0.75 (0.72--0.78) & 0.82 (0.80--0.85) & 0.61 (0.58--0.64) & 0.83 (0.81--0.85) \\
\rowcolor{rowshade}068 & DR grade (RetinaMNIST) & 0.65 (0.61--0.69) & 0.70 (0.66--0.74) & 0.57 (0.54--0.61) & 0.86 (0.83--0.89) \\
069 & Retinal OCT diagnosis & 0.97 (0.97--0.98) & 0.98 (0.98--0.99) & 0.92 (0.90--0.93) & 1.00 (1.00--1.00) \\
\rowcolor{rowshade}070 & Colorectal tissue type & 0.98 (0.97--0.98) & 0.96 (0.95--0.97) & 0.87 (0.86--0.89) & 0.96 (0.95--0.97) \\
071 & Blood cell type & 0.97 (0.97--0.98) & 0.98 (0.98--0.98) & 0.86 (0.85--0.87) & 0.99 (0.98--0.99) \\
\rowcolor{rowshade}072 & ECG superclasses & 0.73 (0.72--0.75) & 0.67 (0.66--0.68) & 0.74 (0.73--0.75) & 0.68 (0.66--0.69) \\
073 & VQA-RAD yes/no & 0.91 (0.87--0.94) & 0.88 (0.84--0.92) & 0.87 (0.82--0.91) & 0.94 (0.91--0.96) \\
\rowcolor{rowshade}074 & SLAKE yes/no & 0.94 (0.92--0.96) & 0.92 (0.89--0.95) & 0.90 (0.87--0.94) & 0.97 (0.96--0.99) \\
075 & SLAKE yes/no (Chinese) & 0.96 (0.94--0.98) & 0.92 (0.89--0.95) & 0.93 (0.90--0.96) & 0.99 (0.98--0.99) \\
\rowcolor{rowshade}076 & PMC-VQA & 0.84 (0.83--0.86) & 0.83 (0.82--0.85) & 0.82 (0.81--0.84) & 0.86 (0.85--0.88) \\
077 & MedXpertQA multimodal & 0.72 (0.70--0.73) & 0.64 (0.63--0.66) & 0.79 (0.78--0.81) & 0.79 (0.78--0.81) \\
\rowcolor{rowshade}078 & Head CT haemorrhage & 0.83 (0.75--0.89) & 0.85 (0.80--0.91) & 0.90 (0.85--0.94) & 0.89 (0.84--0.93) \\
079 & Critical head CT finding & 0.84 (0.78--0.88) & 0.79 (0.74--0.85) & 0.88 (0.83--0.92) & 0.87 (0.83--0.90) \\
\rowcolor{rowshade}080 & Chest CT abnormalities & 0.69 (0.66--0.73) & 0.71 (0.68--0.74) & 0.76 (0.73--0.79) & 0.74 (0.71--0.77) \\
081 & Pulmonary embolism on CTPA & 0.51 (0.45--0.56) & 0.52 (0.45--0.59) & 0.54 (0.49--0.59) & 0.54 (0.49--0.59) \\
\rowcolor{rowshade}082 & Cervical spine fracture & 0.57 (0.51--0.63) & 0.58 (0.53--0.64) & 0.59 (0.54--0.65) & 0.53 (0.47--0.60) \\
083 & Abdominal trauma on CT & 0.69 (0.64--0.74) & 0.73 (0.68--0.77) & 0.74 (0.68--0.79) & 0.79 (0.75--0.83) \\
\rowcolor{rowshade}084 & Pancreatic cancer on CT & 0.61 (0.55--0.68) & 0.68 (0.61--0.74) & 0.64 (0.58--0.71) & 0.70 (0.64--0.75) \\
085 & Prostate cancer bpMRI & 0.58 (0.50--0.66) & 0.44 (0.36--0.51) & 0.55 (0.46--0.63) & 0.59 (0.51--0.67) \\
\rowcolor{rowshade}086 & Rib fracture type & 0.41 (0.37--0.46) & 0.48 (0.43--0.53) & 0.42 (0.37--0.47) & 0.44 (0.39--0.49) \\
\bottomrule
\end{longtable}
}

\clearpage
{\footnotesize
\begin{longtable}{lrlccccc}
\caption[Random-effects means per capability tier]{Random-effects mean per capability tier with 95\% CI. Best denotes the best of the four models on each task.}\label{tab:tiers}\\
\toprule
Tier & Tasks & Metric & Clef & Clef-flash & GPT-6 Luna Decisions & Jev 1.14 & Best \\
\midrule
\endfirsthead
\toprule
Tier & Tasks & Metric & Clef & Clef-flash & GPT-6 Luna Decisions & Jev 1.14 & Best \\
\midrule
\endhead
Medical knowledge & 16 & AUROC & 0.90 (0.83--0.96) & 0.88 (0.81--0.94) & 0.94 (0.91--0.96) & 0.93 (0.89--0.97) & 0.94 (0.92--0.97) \\
 &  & Chance-corrected BA & 0.65 (0.47--0.83) & 0.60 (0.44--0.75) & 0.73 (0.63--0.84) & 0.72 (0.57--0.87) & 0.74 (0.63--0.85) \\
\rowcolor{rowshade}Text understanding & 13 & AUROC & 0.92 (0.90--0.93) & 0.89 (0.87--0.92) & 0.89 (0.87--0.92) & 0.92 (0.90--0.94) & 0.93 (0.91--0.94) \\
\rowcolor{rowshade} &  & Chance-corrected BA & 0.63 (0.54--0.73) & 0.56 (0.43--0.69) & 0.61 (0.51--0.71) & 0.65 (0.55--0.75) & 0.66 (0.57--0.75) \\
Clinical judgement from text & 6 & AUROC & 0.79 (0.70--0.87) & 0.73 (0.64--0.81) & 0.80 (0.66--0.94) & 0.82 (0.73--0.90) & 0.83 (0.73--0.93) \\
 &  & Chance-corrected BA & 0.34 (0.14--0.55) & 0.24 (0.10--0.38) & 0.41 (0.10--0.73) & 0.42 (0.12--0.72) & 0.44 (0.12--0.77) \\
\rowcolor{rowshade}Image perception & 12 & AUROC & 0.84 (0.77--0.92) & 0.85 (0.79--0.91) & 0.85 (0.76--0.93) & 0.88 (0.83--0.94) & 0.89 (0.84--0.94) \\
\rowcolor{rowshade} &  & Chance-corrected BA & 0.38 (0.31--0.45) & 0.40 (0.31--0.48) & 0.41 (0.24--0.59) & 0.46 (0.26--0.66) & 0.45 (0.28--0.61) \\
2D image findings & 22 & AUROC & 0.80 (0.75--0.84) & 0.80 (0.75--0.84) & 0.77 (0.73--0.80) & 0.83 (0.79--0.86) & 0.84 (0.81--0.87) \\
 &  & Chance-corrected BA & 0.28 (0.18--0.38) & 0.29 (0.18--0.40) & 0.26 (0.16--0.35) & 0.39 (0.24--0.53) & 0.40 (0.27--0.53) \\
\rowcolor{rowshade}Volumetric findings & 9 & AUROC & 0.64 (0.55--0.73) & 0.64 (0.56--0.73) & 0.67 (0.56--0.78) & 0.68 (0.58--0.78) & 0.69 (0.60--0.79) \\
\rowcolor{rowshade} &  & Chance-corrected BA & 0.13 (0.05--0.21) & 0.14 (0.04--0.23) & 0.14 (0.07--0.22) & 0.16 (0.04--0.27) & 0.20 (0.09--0.32) \\
\bottomrule
\end{longtable}
}

\clearpage
{\footnotesize
\begin{longtable}{llllrrrr}
\caption[System 1 and generative models]{System 1 model against the generative model of its family on the same cases, 100 per task and 50 for Clef and Qwen3.8-27B. BA denotes balanced accuracy, USD/1k the cost in US dollars per 1,000 requests, S1 the System 1 model and gen. the generative model.}\label{tab:generative}\\
\toprule
No. & Task & System 1 & Generative & BA S1 & BA gen. & USD/1k S1 & USD/1k gen. \\
\midrule
\endfirsthead
\toprule
No. & Task & System 1 & Generative & BA S1 & BA gen. & USD/1k S1 & USD/1k gen. \\
\midrule
\endhead
001 & MedQA (USMLE) & Clef & Qwen3.8-27B & 0.84 & 0.92 & 0.08 & 1.40 \\
\rowcolor{rowshade}027 & Lung cancer T from report & Clef & Qwen3.8-27B & 0.65 & 0.83 & 0.11 & 1.62 \\
028 & BI-RADS from report & Clef & Qwen3.8-27B & 0.54 & 0.35 & 0.12 & 1.64 \\
\rowcolor{rowshade}029 & pT stage from pathology report & Clef & Qwen3.8-27B & 0.85 & 0.86 & 0.32 & 0.96 \\
030 & Trial report inference & Clef & Qwen3.8-27B & 0.90 & 0.89 & 0.15 & 1.27 \\
\rowcolor{rowshade}031 & PubMedQA & Clef & Qwen3.8-27B & 0.57 & 0.70 & 0.12 & 1.03 \\
032 & Drug-drug interaction & Clef & Qwen3.8-27B & 0.80 & 0.88 & 0.08 & 1.15 \\
\rowcolor{rowshade}036 & Rubric judging & Clef & Qwen3.8-27B & 0.82 & 0.64 & 0.26 & 1.70 \\
037 & ED triage (KTAS) & Clef & Qwen3.8-27B & 0.52 & 0.54 & 0.08 & 2.11 \\
\rowcolor{rowshade}038 & Diagnosis from case & Clef & Qwen3.8-27B & 0.55 & 0.61 & 0.14 & 2.21 \\
039 & ACMG variant class & Clef & Qwen3.8-27B & 0.39 & 0.72 & 0.08 & 2.45 \\
\rowcolor{rowshade}040 & Trial eligibility & Clef & Qwen3.8-27B & 0.65 & 0.66 & 0.24 & 1.69 \\
041 & Suicide-risk level & Clef & Qwen3.8-27B & 0.48 & 0.62 & 0.44 & 1.82 \\
\rowcolor{rowshade}042 & MRI safety of a device & Clef & Qwen3.8-27B & 0.25 & 0.25 & 0.07 & 0.72 \\
043 & Radiograph body part & Clef & Qwen3.8-27B & 0.77 & 0.77 & 0.27 & 0.74 \\
\rowcolor{rowshade}045 & Imaging modality & Clef & Qwen3.8-27B & 0.97 & 0.97 & 0.17 & 0.53 \\
046 & Mammography view & Clef & Qwen3.8-27B & 0.61 & 0.81 & 0.19 & 1.73 \\
\rowcolor{rowshade}048 & CT contrast phase (slice) & Clef & Qwen3.8-27B & 0.52 & 0.53 & 0.15 & 2.21 \\
052 & Vertebra level & Clef & Qwen3.8-27B & 0.62 & 0.58 & 0.36 & 2.02 \\
\rowcolor{rowshade}054 & Low-dose CT quality & Clef & Qwen3.8-27B & 0.33 & 0.41 & 0.11 & 1.34 \\
055 & CXR findings & Clef & Qwen3.8-27B & 0.62 & 0.60 & 0.47 & 2.13 \\
\rowcolor{rowshade}057 & Line and tube position & Clef & Qwen3.8-27B & 0.57 & 0.63 & 0.62 & 3.07 \\
058 & Bone-age band & Clef & Qwen3.8-27B & 0.17 & 0.26 & 0.26 & 2.33 \\
\rowcolor{rowshade}059 & Knee OA KL grade & Clef & Qwen3.8-27B & 0.20 & 0.17 & 0.09 & 2.24 \\
060 & Lumbar stenosis grade & Clef & Qwen3.8-27B & 0.43 & 0.35 & 0.50 & 2.92 \\
\rowcolor{rowshade}061 & Head CT haemorrhage (slice) & Clef & Qwen3.8-27B & 0.56 & 0.56 & 0.29 & 2.19 \\
062 & BI-RADS from mammogram & Clef & Qwen3.8-27B & 0.27 & 0.31 & 0.26 & 2.41 \\
\rowcolor{rowshade}063 & Breast density & Clef & Qwen3.8-27B & 0.28 & 0.28 & 0.21 & 1.95 \\
064 & Dermoscopy (HAM10000) & Clef & Qwen3.8-27B & 0.47 & 0.46 & 0.14 & 1.83 \\
\rowcolor{rowshade}067 & DR severity (IDRiD) & Clef & Qwen3.8-27B & 0.30 & 0.30 & 0.24 & 1.57 \\
070 & Colorectal tissue type & Clef & Qwen3.8-27B & 0.79 & 0.75 & 0.09 & 1.59 \\
\rowcolor{rowshade}072 & ECG superclasses & Clef & Qwen3.8-27B & 0.49 & 0.62 & 0.38 & 3.11 \\
001 & MedQA (USMLE) & Clef-flash & Qwen3.5-9B & 0.72 & 0.92 & 0.01 & 0.65 \\
\rowcolor{rowshade}027 & Lung cancer T from report & Clef-flash & Qwen3.5-9B & 0.48 & 0.84 & 0.02 & 0.81 \\
028 & BI-RADS from report & Clef-flash & Qwen3.5-9B & 0.51 & 0.32 & 0.02 & 0.80 \\
\rowcolor{rowshade}029 & pT stage from pathology report & Clef-flash & Qwen3.5-9B & 0.79 & 0.89 & 0.05 & 0.61 \\
030 & Trial report inference & Clef-flash & Qwen3.5-9B & 0.79 & 0.90 & 0.03 & 0.72 \\
\rowcolor{rowshade}031 & PubMedQA & Clef-flash & Qwen3.5-9B & 0.57 & 0.58 & 0.02 & 0.71 \\
032 & Drug-drug interaction & Clef-flash & Qwen3.5-9B & 0.69 & 0.80 & 0.01 & 0.78 \\
\rowcolor{rowshade}036 & Rubric judging & Clef-flash & Qwen3.5-9B & 0.65 & 0.65 & 0.04 & 0.72 \\
037 & ED triage (KTAS) & Clef-flash & Qwen3.5-9B & 0.46 & 0.47 & 0.01 & 0.89 \\
\rowcolor{rowshade}038 & Diagnosis from case & Clef-flash & Qwen3.5-9B & 0.45 & 0.62 & 0.02 & 0.87 \\
039 & ACMG variant class & Clef-flash & Qwen3.5-9B & 0.34 & 0.64 & 0.01 & 1.17 \\
\rowcolor{rowshade}040 & Trial eligibility & Clef-flash & Qwen3.5-9B & 0.67 & 0.68 & 0.04 & 0.83 \\
041 & Suicide-risk level & Clef-flash & Qwen3.5-9B & 0.28 & 0.53 & 0.07 & 1.00 \\
\rowcolor{rowshade}042 & MRI safety of a device & Clef-flash & Qwen3.5-9B & 0.21 & 0.23 & 0.01 & 0.88 \\
043 & Radiograph body part & Clef-flash & Qwen3.5-9B & 0.84 & 0.81 & 0.04 & 0.23 \\
\rowcolor{rowshade}045 & Imaging modality & Clef-flash & Qwen3.5-9B & 0.97 & 0.97 & 0.03 & 0.19 \\
046 & Mammography view & Clef-flash & Qwen3.5-9B & 0.68 & 0.74 & 0.03 & 0.29 \\
\rowcolor{rowshade}048 & CT contrast phase (slice) & Clef-flash & Qwen3.5-9B & 0.64 & 0.58 & 0.02 & 0.37 \\
052 & Vertebra level & Clef-flash & Qwen3.5-9B & 0.53 & 0.47 & 0.06 & 1.27 \\
\rowcolor{rowshade}054 & Low-dose CT quality & Clef-flash & Qwen3.5-9B & 0.33 & 0.37 & 0.02 & 0.23 \\
055 & CXR findings & Clef-flash & Qwen3.5-9B & 0.65 & 0.65 & 0.08 & 0.72 \\
\rowcolor{rowshade}057 & Line and tube position & Clef-flash & Qwen3.5-9B & 0.58 & 0.60 & 0.10 & 1.27 \\
058 & Bone-age band & Clef-flash & Qwen3.5-9B & 0.23 & 0.24 & 0.04 & 1.50 \\
\rowcolor{rowshade}059 & Knee OA KL grade & Clef-flash & Qwen3.5-9B & 0.35 & 0.28 & 0.01 & 0.39 \\
060 & Lumbar stenosis grade & Clef-flash & Qwen3.5-9B & 0.33 & 0.39 & 0.08 & 1.84 \\
\rowcolor{rowshade}061 & Head CT haemorrhage (slice) & Clef-flash & Qwen3.5-9B & 0.56 & 0.62 & 0.05 & 0.61 \\
062 & BI-RADS from mammogram & Clef-flash & Qwen3.5-9B & 0.33 & 0.37 & 0.04 & 1.04 \\
\rowcolor{rowshade}063 & Breast density & Clef-flash & Qwen3.5-9B & 0.33 & 0.30 & 0.03 & 0.44 \\
064 & Dermoscopy (HAM10000) & Clef-flash & Qwen3.5-9B & 0.52 & 0.53 & 0.02 & 0.49 \\
\rowcolor{rowshade}067 & DR severity (IDRiD) & Clef-flash & Qwen3.5-9B & 0.32 & 0.35 & 0.04 & 0.83 \\
070 & Colorectal tissue type & Clef-flash & Qwen3.5-9B & 0.65 & 0.68 & 0.01 & 0.25 \\
\rowcolor{rowshade}072 & ECG superclasses & Clef-flash & Qwen3.5-9B & 0.51 & 0.51 & 0.06 & 1.75 \\
001 & MedQA (USMLE) & GPT-6 Luna Decisions & GPT-6 Luna & 0.90 & 0.97 & 0.03 & 0.08 \\
\rowcolor{rowshade}027 & Lung cancer T from report & GPT-6 Luna Decisions & GPT-6 Luna & 0.76 & 0.91 & 0.04 & 0.09 \\
028 & BI-RADS from report & GPT-6 Luna Decisions & GPT-6 Luna & 0.32 & 0.37 & 0.04 & 0.09 \\
\rowcolor{rowshade}029 & pT stage from pathology report & GPT-6 Luna Decisions & GPT-6 Luna & 0.86 & 0.87 & 0.13 & 0.18 \\
030 & Trial report inference & GPT-6 Luna Decisions & GPT-6 Luna & 0.79 & 0.86 & 0.07 & 0.12 \\
\rowcolor{rowshade}031 & PubMedQA & GPT-6 Luna Decisions & GPT-6 Luna & 0.55 & 0.63 & 0.05 & 0.08 \\
032 & Drug-drug interaction & GPT-6 Luna Decisions & GPT-6 Luna & 0.79 & 0.84 & 0.03 & 0.06 \\
\rowcolor{rowshade}036 & Rubric judging & GPT-6 Luna Decisions & GPT-6 Luna & 0.64 & 0.62 & 0.11 & 0.20 \\
037 & ED triage (KTAS) & GPT-6 Luna Decisions & GPT-6 Luna & 0.53 & 0.56 & 0.03 & 0.10 \\
\rowcolor{rowshade}038 & Diagnosis from case & GPT-6 Luna Decisions & GPT-6 Luna & 0.66 & 0.77 & 0.05 & 0.12 \\
039 & ACMG variant class & GPT-6 Luna Decisions & GPT-6 Luna & 0.85 & 0.82 & 0.03 & 0.13 \\
\rowcolor{rowshade}040 & Trial eligibility & GPT-6 Luna Decisions & GPT-6 Luna & 0.64 & 0.60 & 0.09 & 0.15 \\
041 & Suicide-risk level & GPT-6 Luna Decisions & GPT-6 Luna & 0.54 & 0.61 & 0.19 & 0.27 \\
\rowcolor{rowshade}042 & MRI safety of a device & GPT-6 Luna Decisions & GPT-6 Luna & 0.25 & 0.30 & 0.02 & 0.12 \\
043 & Radiograph body part & GPT-6 Luna Decisions & GPT-6 Luna & 0.88 & 0.90 & 0.11 & 0.18 \\
\rowcolor{rowshade}045 & Imaging modality & GPT-6 Luna Decisions & GPT-6 Luna & 0.96 & 0.95 & 0.07 & 0.09 \\
046 & Mammography view & GPT-6 Luna Decisions & GPT-6 Luna & 0.74 & 0.79 & 0.08 & 0.13 \\
\rowcolor{rowshade}048 & CT contrast phase (slice) & GPT-6 Luna Decisions & GPT-6 Luna & 0.72 & 0.79 & 0.07 & 0.16 \\
052 & Vertebra level & GPT-6 Luna Decisions & GPT-6 Luna & 0.64 & 0.58 & 0.13 & 0.60 \\
\rowcolor{rowshade}054 & Low-dose CT quality & GPT-6 Luna Decisions & GPT-6 Luna & 0.40 & 0.46 & 0.04 & 0.08 \\
055 & CXR findings & GPT-6 Luna Decisions & GPT-6 Luna & 0.67 & 0.65 & 0.29 & 0.38 \\
\rowcolor{rowshade}057 & Line and tube position & GPT-6 Luna Decisions & GPT-6 Luna & 0.55 & 0.62 & 0.32 & 0.79 \\
058 & Bone-age band & GPT-6 Luna Decisions & GPT-6 Luna & 0.34 & 0.45 & 0.10 & 0.30 \\
\rowcolor{rowshade}059 & Knee OA KL grade & GPT-6 Luna Decisions & GPT-6 Luna & 0.25 & 0.33 & 0.03 & 0.13 \\
060 & Lumbar stenosis grade & GPT-6 Luna Decisions & GPT-6 Luna & 0.50 & 0.57 & 0.21 & 0.64 \\
\rowcolor{rowshade}061 & Head CT haemorrhage (slice) & GPT-6 Luna Decisions & GPT-6 Luna & 0.59 & 0.60 & 0.15 & 0.26 \\
062 & BI-RADS from mammogram & GPT-6 Luna Decisions & GPT-6 Luna & 0.29 & 0.44 & 0.10 & 0.27 \\
\rowcolor{rowshade}063 & Breast density & GPT-6 Luna Decisions & GPT-6 Luna & 0.33 & 0.32 & 0.08 & 0.14 \\
064 & Dermoscopy (HAM10000) & GPT-6 Luna Decisions & GPT-6 Luna & 0.44 & 0.51 & 0.05 & 0.16 \\
\rowcolor{rowshade}067 & DR severity (IDRiD) & GPT-6 Luna Decisions & GPT-6 Luna & 0.33 & 0.42 & 0.10 & 0.22 \\
070 & Colorectal tissue type & GPT-6 Luna Decisions & GPT-6 Luna & 0.55 & 0.45 & 0.03 & 0.10 \\
\rowcolor{rowshade}072 & ECG superclasses & GPT-6 Luna Decisions & GPT-6 Luna & 0.57 & 0.62 & 0.19 & 0.75 \\
\bottomrule
\end{longtable}
}

\clearpage
{\footnotesize
\begin{longtable}{lllP{7.4cm}P{3.4cm}lrr}
\caption[Published results]{Published results with the same dataset, label set and metric, which can use other splits, and the best of the four System 1 models on the questions that match the published result (Best S1). Best S1 can therefore differ from the task-level AUROC in Table~\ref{tab:per_task}.}\label{tab:references}\\
\toprule
No. & Task & Type & Reference & Source & Metric & Ref. & Best S1 \\
\midrule
\endfirsthead
\toprule
No. & Task & Type & Reference & Source & Metric & Ref. & Best S1 \\
\midrule
\endhead
001 & MedQA (USMLE) & Generative language model & o1-preview, zero-shot & Nori 2024 & accuracy & 0.96 & 0.89 \\
\rowcolor{rowshade}001 & MedQA (USMLE) & Task-specific model & Med-Gemini-L 1.0 (self-training, uncertainty-guided web search) & Saab 2024 & accuracy & 0.91 & 0.89 \\
002 & MedMCQA (India) & Generative language model & o1-preview, zero-shot & Nori 2024 & accuracy & 0.84 & 0.75 \\
\rowcolor{rowshade}002 & MedMCQA (India) & Task-specific model & Med-PaLM 2 (ensemble refinement) & Singhal 2025 & accuracy & 0.72 & 0.75 \\
003 & HEAD-QA (English) & Generative language model & GPT-5, zero-shot & Yang 2026 & accuracy & 0.90 & 0.90 \\
\rowcolor{rowshade}005 & MMLU medical & Generative language model & GPT-5 & Wang 2025 & accuracy & 0.96 & 0.96 \\
006 & MMLU-Pro health & Generative language model & Gemini-3.1-Pro & Wang 2024 (leaderboard) & accuracy & 0.86 & 0.83 \\
\rowcolor{rowshade}007 & MedXpertQA text & Clinicians & Expert (pre-licensed humans) & Zuo 2025 & accuracy & 0.43 & 0.35 \\
007 & MedXpertQA text & Generative language model & GPT-5 & Wang 2025 & accuracy & 0.56 & 0.35 \\
\rowcolor{rowshade}008 & Medbullets & Generative language model & GPT-5 (2025-08-07) & Bedi 2025 (MedHELM) & accuracy & 0.87 & 0.78 \\
009 & AfriMed-QA & Generative language model & GPT-4o, zero-shot & Olatunji 2025 & accuracy & 0.79 & 0.82 \\
\rowcolor{rowshade}010 & MedExpQA & Generative language model & o4-mini, zero-shot & Yang 2026 & accuracy & 0.88 & 0.87 \\
010 & MedExpQA & Task-specific model & Mistral-7B (fine-tuned), no RAG & Alonso 2024 & accuracy & 0.68 & 0.87 \\
\rowcolor{rowshade}011 & KorMedMCQA & Clinicians & Examinees average (licensing exam institute) & Kweon 2024 & accuracy & 0.77 & 0.89 \\
011 & KorMedMCQA & Generative language model & o1-preview, 5-shot & Kweon 2024 & accuracy & 0.93 & 0.89 \\
\rowcolor{rowshade}012 & CMExam (China) & Clinicians & Human exam takers & Liu 2023 & accuracy & 0.72 & 0.88 \\
012 & CMExam (China) & Generative language model & Qwen2.5-72B-Instruct & Baichuan 2025 & accuracy & 0.83 & 0.88 \\
\rowcolor{rowshade}012 & CMExam (China) & Task-specific model & ChatGLM-CMExam (fine-tuned) & Liu 2023 & accuracy & 0.45 & 0.88 \\
013 & MedQA Mainland & Generative language model & GPT-4 (5-shot CoT) & Qiu 2024 & accuracy & 0.81 & 0.90 \\
\rowcolor{rowshade}013 & MedQA Mainland & Task-specific model & MMed-Llama 3 8B (fine-tuned) & Qiu 2024 & accuracy & 0.79 & 0.90 \\
014 & MedQA Taiwan & Generative language model & GPT-4-base (pre-alignment), 5-shot & Nori 2023 & accuracy & 0.87 & 0.90 \\
\rowcolor{rowshade}014 & MedQA Taiwan & Task-specific model & BioBERT-large (machine-translated to English + retrieval) & Jin 2021 & accuracy & 0.42 & 0.90 \\
015 & FrenchMedMCQA & Generative language model & GPT-4, 3-shot & Wang 2024 (Apollo) & accuracy & 0.90 & 0.93 \\
\rowcolor{rowshade}022 & ATC drug class & Generative language model & GPT-4-0125-preview, 4-shot & Shoham 2024 & accuracy & 0.62 & 0.62 \\
023 & ICD-9 procedure & Generative language model & GPT-4-0125-preview, 4-shot & Shoham 2024 & accuracy & 0.55 & 0.53 \\
\rowcolor{rowshade}025 & Chest CT report findings & Generative language model & GPT-4o (gpt-4o-2024-11-20), zero-shot, JSON output & Yamagishi 2025 & accuracy & 0.97 & 0.96 \\
027 & Lung cancer T from report & Generative language model & GPT-3.5-turbo, zero-shot, full TNM definitions in prompt & Matsuo 2024 & accuracy & 0.57 & 0.78 \\
\rowcolor{rowshade}027 & Lung cancer T from report & Task-specific model & NAISTSOCRR-2 (fine-tuned Japanese BERTs) & Nakamura 2023 & accuracy & 0.67 & 0.78 \\
029 & pT stage from pathology report & Task-specific model & BB-TEN (Clinical-BigBird, fine-tuned) & Kefeli 2024 & AUROC & 0.94 & 0.96 \\
\rowcolor{rowshade}031 & PubMedQA & Clinicians & Single human annotator & Jin 2019 & accuracy & 0.78 & 0.78 \\
031 & PubMedQA & Generative language model & GPT-4 + Medprompt (kNN few-shot, self-generated CoT, choice-shuffle ensemble) & Nori 2023 (Medprompt) & accuracy & 0.82 & 0.78 \\
\rowcolor{rowshade}031 & PubMedQA & Task-specific model & MedGemma 27B (text, test-time scaling) & Sellergren 2025 & accuracy & 0.77 & 0.78 \\
035 & Question entailment & Generative language model & gemma-4-31B-it, five-shot & Wu 2025 (BRIDGE) & accuracy & 0.88 & 0.83 \\
\rowcolor{rowshade}035 & Question entailment & Task-specific model & PANLP (ensemble, rank 1) & Ben Abacha 2019 & accuracy & 0.75 & 0.83 \\
037 & ED triage (KTAS) & Clinicians & Triage nurses (KTAS\_RN) & Moon 2019 (in-data) & accuracy & 0.85 & 0.56 \\
\rowcolor{rowshade}037 & ED triage (KTAS) & Generative language model & KATE (GPT-4o, 40 kNN shots) & Lee 2026 & accuracy & 0.72 & 0.56 \\
037 & ED triage (KTAS) & Task-specific model & Llama 3.1 8B, fine-tuned & Lee 2026 & accuracy & 0.65 & 0.56 \\
\rowcolor{rowshade}038 & Diagnosis from case & Generative language model & o3 & Zhu 2026 & accuracy & 0.69 & 0.65 \\
039 & ACMG variant class & Rule & ACMG/AMP points system on panel evidence codes & Tavtigian 2020 (in-data) & accuracy & 0.95 & 0.73 \\
\rowcolor{rowshade}040 & Trial eligibility & Generative language model & TrialGPT-Ranking (GPT-4) & Jin 2024 & AUROC & 0.80 & 0.87 \\
040 & Trial eligibility & Task-specific model & BioLinkBERT cross-encoder (MedNLI) & Jin 2024 & AUROC & 0.62 & 0.87 \\
\rowcolor{rowshade}041 & Suicide-risk level & Generative language model & GPT-4, zero-shot (best prompt) & Xu 2024 (Mental-LLM) & balanced accuracy & 0.44 & 0.56 \\
041 & Suicide-risk level & Task-specific model & Mental-FLAN-T5 (FLAN-T5-XXL, instruction-finetuned on four mental-health datasets) & Xu 2024 (Mental-LLM) & balanced accuracy & 0.48 & 0.56 \\
\rowcolor{rowshade}044 & CXR view AP/PA & Task-specific model & ResNet-18 & Kim 2019 & AUROC & 1.00 & 0.99 \\
046 & Mammography view & Task-specific model & ResNet-50, fine-tuned, 256 × 256 px & Yi 2019 & AUROC & 1.00 & 0.99 \\
\rowcolor{rowshade}047 & CT contrast phase (series) & Task-specific model & 2dFM\_BERT (frozen MAE ViT-L/16 + BERT) & Hou 2025 & AUROC & 0.91 & 0.90 \\
049 & Abdominal organ (3D) & Task-specific model & 3D CNN with Kinetics-400 video initialization, fine-tuned & Jain 2024 & accuracy & 0.97 & 0.38 \\
\rowcolor{rowshade}050 & Abdominal organ (coronal) & Generative language model & Qwen3-VL-8B with retrieved exemplars and rules (in-context learning, frozen) & Zhao 2026 & accuracy & 0.94 & 0.40 \\
050 & Abdominal organ (coronal) & Task-specific model & DenseNet-121, 128 px & Doerrich 2024 & accuracy & 0.94 & 0.40 \\
\rowcolor{rowshade}051 & Abdominal organ (sagittal) & Task-specific model & DenseNet-121, 64 px & Doerrich 2024 & accuracy & 0.83 & 0.31 \\
053 & MRI motion artefact & Task-specific model & FewSOME (unsupervised few-shot) & Belton 2024 & accuracy & 0.72 & 0.32 \\
\rowcolor{rowshade}055 & CXR findings & Task-specific model & Ark+ (Swin-L, 768 px; fine-tuned) & Ma 2025 (Ark+) & AUROC & 0.84 & 0.77 \\
056 & Paediatric pneumonia & Generative language model & MedGemma-27B, evolved prompt, no fine-tuning & Sviridov 2026 & accuracy & 0.84 & 0.79 \\
\rowcolor{rowshade}056 & Paediatric pneumonia & Task-specific model & DINO ViT-B/16, linear probing, 224 px & Doerrich 2025 & AUROC & 0.99 & 0.96 \\
057 & Line and tube position & Clinicians & Radiologist (Reader 2 of 2, board-certified/board-eligible) & Güzel 2026 & balanced accuracy & 0.75 & 0.50 \\
\rowcolor{rowshade}057 & Line and tube position & Generative language model & Gemini 3.1 Flash Lite Preview, zero-shot (Swan-Ganz) & Güzel 2026 & balanced accuracy & 0.54 & 0.87 \\
057 & Line and tube position & Task-specific model & Rank 1 (All Data Are Ext) & Kaggle RANZCR CLiP private LB & AUROC & 0.98 & 0.81 \\
\rowcolor{rowshade}059 & Knee OA KL grade & Task-specific model & Ensemble of 6 ImageNet-pretrained CNNs, per-model optimised input size, mix voting & Pi 2023 & accuracy & 0.77 & 0.42 \\
061 & Head CT haemorrhage (slice) & Generative language model & Gemini 2.0 Flash, zero-shot & Wang 2025 & accuracy & 0.75 & 0.71 \\
\rowcolor{rowshade}061 & Head CT haemorrhage (slice) & Task-specific model & Rank-1 Kaggle ensemble & Wang 2021 & AUROC & 1.00 & 0.88 \\
063 & Breast density & Task-specific model & LG-DPTNet: multi-view (CC + MLO) fusion, dynamic pyramid convolution + cross-transformer & Zhong 2023 & AUROC & 0.97 & 0.71 \\
\rowcolor{rowshade}064 & Dermoscopy (HAM10000) & Clinicians & Human experts (> 10 years) & Tschandl 2019 & balanced accuracy & 0.81 & 0.62 \\
064 & Dermoscopy (HAM10000) & Generative language model & Hulu-Med-7B, zero-shot & Liu 2026 (DermAgent) & accuracy & 0.53 & 0.63 \\
\rowcolor{rowshade}064 & Dermoscopy (HAM10000) & Task-specific model & Rank 1, team 'Minjie', approach 'ensemble b4 ce+fl+cepos+ds1 b3 ce' & ISIC 2018 live leaderboard & balanced accuracy & 0.90 & 0.62 \\
065 & Dermoscopy (ISIC 2019) & Task-specific model & Rank 1 (DAISYLab, EfficientNet ensemble, external data) & ISIC 2019 leaderboard & balanced accuracy & 0.64 & 0.48 \\
\rowcolor{rowshade}066 & Melanoma (ISIC 2020) & Task-specific model & Rank 1 of 3,308 entries (CNN ensemble) & Kurtansky 2024 & AUROC & 0.95 & 0.78 \\
067 & DR severity (IDRiD) & Task-specific model & Multi-stage transfer learning (ImageNet $\rightarrow$ EyePACS $\rightarrow$ DDR $\rightarrow$ IDRiD) + class-balanced loss & Shi 2023 & accuracy & 0.80 & 0.56 \\
\rowcolor{rowshade}068 & DR grade (RetinaMNIST) & Generative language model & Gemini-3-flash, zero-shot & Zhao 2026 & accuracy & 0.44 & 0.66 \\
068 & DR grade (RetinaMNIST) & Task-specific model & VGG16, end-to-end training, 224 × 224 & Doerrich 2025 & accuracy & 0.64 & 0.66 \\
\rowcolor{rowshade}069 & Retinal OCT diagnosis & Task-specific model & Inception V3, ImageNet transfer learning (108,312 training images) & Kermany 2018 & accuracy & 0.97 & 0.97 \\
070 & Colorectal tissue type & Generative language model & GPT-4V, 5-shot in-context (kNN) & Ferber 2024 & accuracy & 0.78 & 0.74 \\
\rowcolor{rowshade}070 & Colorectal tissue type & Task-specific model & CONCH, linear probe trained on NCT-CRC-HE-100K & Lu 2024 (CONCH) & balanced accuracy & 0.94 & 0.74 \\
071 & Blood cell type & Task-specific model & DenseNet-121, end-to-end, 224 px & Doerrich 2025 & accuracy & 0.99 & 0.80 \\
\rowcolor{rowshade}072 & ECG superclasses & Generative language model & GPT-4o, zero-shot & Liu 2024 (PULSE) & AUROC & 0.56 & 0.74 \\
072 & ECG superclasses & Task-specific model & Ensemble of all models & Strodthoff 2021 & AUROC & 0.93 & 0.74 \\
\rowcolor{rowshade}073 & VQA-RAD yes/no & Generative language model & GPT-5-mini, zero-shot chain-of-thought & Wang 2025 & accuracy & 0.75 & 0.84 \\
073 & VQA-RAD yes/no & Task-specific model & MedVInT-TD (fine-tuned) & Zhang 2023 & accuracy & 0.87 & 0.84 \\
\rowcolor{rowshade}074 & SLAKE yes/no & Generative language model & MedGemma 4B, zero-shot & Sellergren 2025 & accuracy & 0.88 & 0.93 \\
074 & SLAKE yes/no & Task-specific model & M2I2 (supervised) & Li 2023 & accuracy & 0.91 & 0.93 \\
\rowcolor{rowshade}076 & PMC-VQA & Generative language model & LLaVA-Med, zero-shot & Zhang 2023 & accuracy & 0.35 & 0.64 \\
076 & PMC-VQA & Task-specific model & MedVInT-TD (trained on PMC-VQA) & Zhang 2023 & accuracy & 0.40 & 0.64 \\
\rowcolor{rowshade}077 & MedXpertQA multimodal & Clinicians & Expert (pre-licensed humans) & Zuo 2025 & accuracy & 0.46 & 0.49 \\
077 & MedXpertQA multimodal & Generative language model & GPT-5 & Wang 2025 & accuracy & 0.72 & 0.49 \\
\rowcolor{rowshade}078 & Head CT haemorrhage & Task-specific model & Qure.ai deep learning algorithms & Chilamkurthy 2018 & AUROC & 0.97 & 0.84 \\
079 & Critical head CT finding & Task-specific model & Qure.ai deep learning algorithms & Chilamkurthy 2018 & AUROC & 0.97 & 0.99 \\
\rowcolor{rowshade}080 & Chest CT abnormalities & Task-specific model & CT-CLIP VocabFine & Hamamci 2026 & AUROC & 0.76 & 0.76 \\
081 & Pulmonary embolism on CTPA & Task-specific model & Attention-based CNN-RNN, strongly supervised (all slice labels) & Hu 2025 & AUROC & 0.93 & 0.58 \\
\rowcolor{rowshade}082 & Cervical spine fracture & Task-specific model & Mean of top 8 Kaggle algorithms & Lee 2024 & AUROC & 0.96 & 0.63 \\
083 & Abdominal trauma on CT & Task-specific model & Mean of 8 winning challenge models & Hermans 2025 & AUROC & 0.98 & 0.85 \\
\rowcolor{rowshade}084 & Pancreatic cancer on CT & Clinicians & 68 radiologists (pooled) & Alves 2026 & AUROC & 0.88 & 0.70 \\
084 & Pancreatic cancer on CT & Task-specific model & Rank 1 (MountainHill / DTI, DTI\_PANORAMA) & PANORAMA leaderboard & AUROC & 0.93 & 0.70 \\
\rowcolor{rowshade}085 & Prostate cancer bpMRI & Clinicians & 62 radiologists (mpMRI, PI-RADS 2.1) & Saha 2024 & AUROC & 0.86 & 0.59 \\
085 & Prostate cancer bpMRI & Task-specific model & PI-CAI AI system & Saha 2024 & AUROC & 0.93 & 0.59 \\
\rowcolor{rowshade}086 & Rib fracture type & Task-specific model & Evolutionary neural architecture search with genetic-algorithm data augmentation & Ali 2024 & accuracy & 0.69 & 0.41 \\
\bottomrule
\end{longtable}
}

\end{landscape}

\end{document}